\documentclass[runningheads]{llncs}

\usepackage{eccv}
\usepackage{eccvabbrv}

\usepackage{graphicx}
\usepackage{booktabs}
\usepackage{wrapfig}
\usepackage{marvosym}
\usepackage{colortbl}
\usepackage[most]{tcolorbox}
\usepackage{multirow}
\usepackage{xcolor}
\usepackage{ulem}
\usepackage{xspace}

\definecolor{navyblue}{HTML}{0071BC}
\definecolor{oai-green-200}{HTML}{D9FFD8}
\definecolor{oai-green-400}{HTML}{A6FFA3}
\definecolor{oai-green-600}{HTML}{51DA4C}
\definecolor{oai-gray-300}{HTML}{E5E5E5}
\definecolor{oai-gray-600}{HTML}{A8A8A8}

\newcommand{\bench}{\texttt{FSU-Bench}\xspace}

\newcommand{\FI}{Foresight Intelligence}
\newcommand{\dataset}{\texttt{FSU-QA}\xspace}

\graphicspath{{Main/}{Supplementary_Material/}}

\usepackage{hyperref}
\usepackage{orcidlink}

\begin{document}

% ================================================================
%  PART 1: Main Paper
% ================================================================

% ---------------------------------------------------------------
\title{Thinking Ahead: Foresight Intelligence in MLLMs and World Model}

\titlerunning{Foresight Intelligence in MLLMs and World Model}

\DeclareRobustCommand{\corrauthor}{\textsuperscript{\Letter}}
\DeclareRobustCommand{\equalcont}{\textsuperscript{*}}

\author{Zhantao Gong\inst{1} \and
Liaoyuan Fan\inst{2} \and
Qing Guo\inst{3,4} \and
Xun Xu\inst{5} \and
Xulei Yang\corrauthor\inst{5} \and
Shijie Li\corrauthor\inst{5}}

\authorrunning{Z.~Gong et al.}

\institute{College of Software, Nankai University, China \and
The University of Hong Kong, China \and
NKIARI, Shenzhen Futian, China \and
AAIS \& VCIP, Nankai University, China\and
A*STAR Institute of Advanced Intelligence and Computing, Singapore \\[0.4em]
\email{\{li\_shijie, yang\_xulei\}@a-star.edu.sg}
\url{https://huggingface.co/datasets/Gong-Grant/FSU-QA}}
\maketitle

\begingroup
\renewcommand{\thefootnote}{}
\footnotetext{
* Equal contribution. \qquad
\Letter\ Corresponding author.
}
\endgroup

\begin{abstract}
In this work, we introduce \dataset, a VQA dataset for autonomous driving 
scenarios designed to advance research on Foresight Intelligence---the 
ability to anticipate and reason about complex, long-horizon futures. 
Unlike existing benchmarks that mainly focus on immediate perception or reactive
planning, \dataset evaluates future-oriented driving understanding through
multi-agent-aware and rule-grounded counterfactual QA. 
Rather than holding surrounding agents fixed or targeting geometric path 
generation, our benchmark requires models to infer semantic future outcomes 
from front-view historical observations and past ego trajectories.
A comprehensive evaluation on the accompanying \bench\ reveals that 
state-of-the-art VLMs still face significant challenges in anticipating 
future events. Furthermore, beyond model performance, we examine whether WM-generated predictions remain semantically consistent by using VLM-based proxy judges, and validate this evaluation protocol through shuffled control experiments.
Fine-tuning models on \dataset\ leads to substantial improvements in 
foresight understanding, demonstrating the dataset's effectiveness and 
offering a principled foundation for future research.
\end{abstract}

\section{Introduction}
\label{sec:intro}
In recent years, Vision-Large Language Models (VLMs) have achieved remarkable progress in visual understanding and cross-modal reasoning \cite{liu_visual_2023}. Typically, VLMs integrate a vision encoder with a Large Language Model (LLM) that has been pretrained on large-scale textual corpora rich in common knowledge.
This design enables them to excel in both perception and understanding, leading to their widespread application in domains such as robotic perception and scene understanding \cite{xu_drivegpt4_2024}.

Most existing Vision--Language Model (VLM) studies focus on enabling VLMs to understand multimodal data such as 2D/3D scene understanding, spatial reasoning, and video comprehension~\cite{jain2025unifying,cheng2024spatialrgpt, 11092837}.
However, these works mainly emphasize observed information while neglecting the ability to anticipate and reason about future, unobserved events. This is particularly challenging, as it requires reasoning over complex spatio-temporal information and dealing with multiple plausible futures, which introduces significant uncertainty. We term this concept \FI, which refers to the ability to anticipate and reason about complex future possibilities based on historical observations.

\FI is conceptually distinct from the planning-oriented reasoning studied 
in prior benchmarks such as DriveLM~\cite{sima2024drivelm} and 
OmniDrive~\cite{wang_omnidrive_2025}. Those benchmarks primarily ask models to issue 
immediate control decisions: given the current scene, what should the ego vehicle 
do next? \FI poses a fundamentally different question: \emph{given a hypothetical 
action or event, what are its semantic consequences as the scene unfolds?} This 
requires modeling multi-agent interactions under uncertainty, performing 
counterfactual simulation, and reasoning over extended temporal horizons---tasks 
that are simulation-level in nature rather than reactive.

\begin{figure}[tb]
  \centering
  \includegraphics[width=0.95\linewidth]{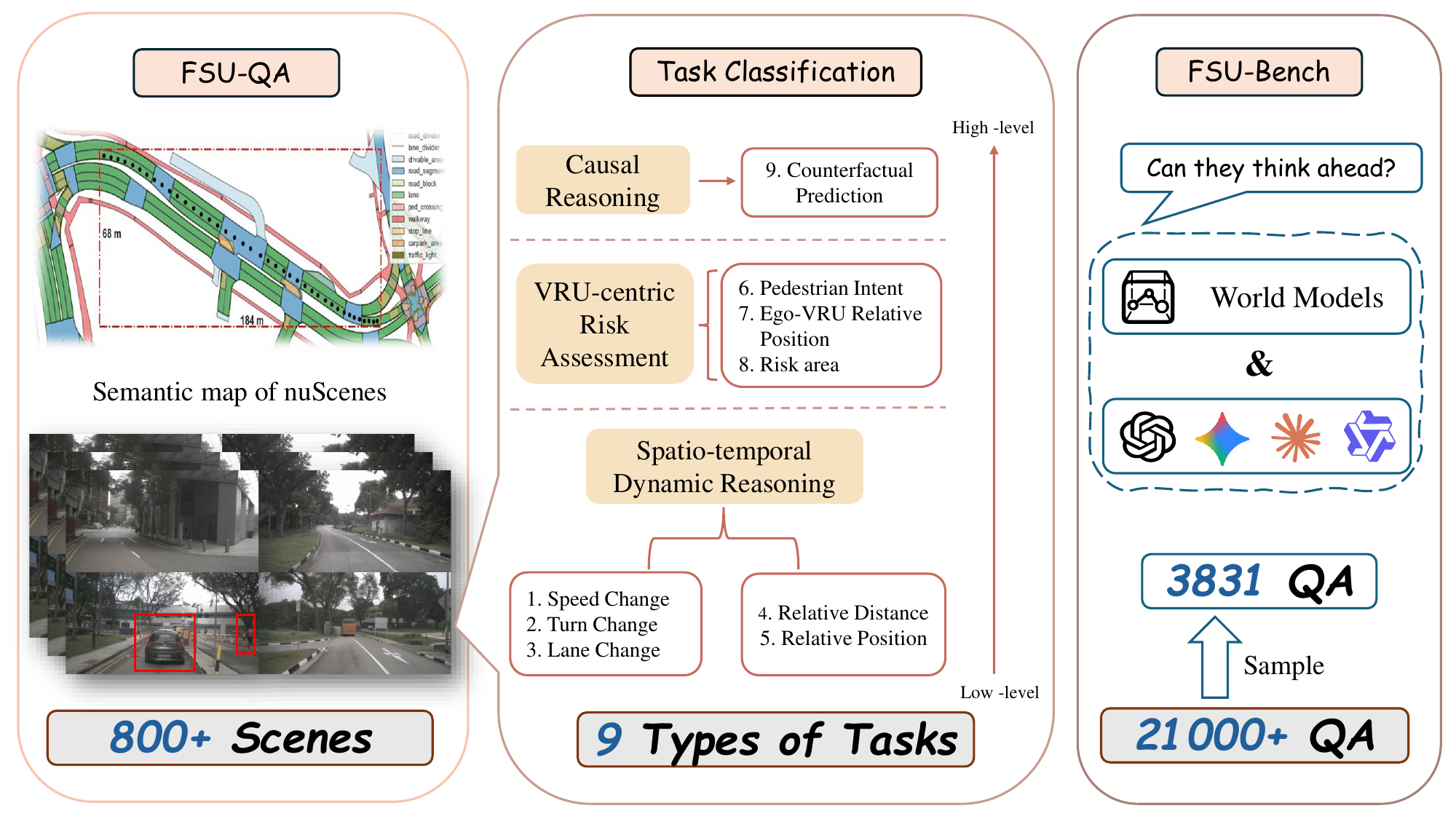}
  \caption{\textbf{Overview of \dataset and \bench.}}
  \label{fig:teaser}
\end{figure}

In this work, we present \dataset (see \cref{fig:teaser}), a Visual Question Answering (VQA) dataset, along with its accompanying benchmark, \bench, designed for training and evaluating Foresight Understanding. Together, they are designed to advance research on \FI.
Considering that VLMs are increasingly being integrated into autonomous driving applications, where vehicles must operate in highly dynamic environments and are required to anticipate potential future events to ensure safety, we construct \dataset under the autonomous driving scenario. In \dataset, we meticulously design the tasks to both train models and comprehensively evaluate their Foresight Intelligence. Specifically, \bench requires the models to anticipate and interpret potential future scenarios across different temporal intervals based on the current situation, thereby enabling effective evaluation of long-horizon foresight capabilities. The designed tasks are organized according to their cognitive complexity, ranging from low-level perception tasks to mid-level imagination tasks and high-level reasoning tasks, which collectively provide a thorough assessment of foresight intelligence.

\bench is also designed to evaluate World Models (WMs)' ability to produce semantically coherent data. While VLMs excel at interpreting observed scenes, WMs incorporate physical 
realism into generative modeling to synthesize plausible future scenarios. Although such data may appear visually realistic, it remains an open question whether they are semantically coherent and can contribute to Foresight Intelligence. To address this, \bench adopts a proxy-evaluation paradigm: VLMs serve as discriminative probes to measure whether WM-generated futures encode semantically meaningful predictive content, quantified by downstream QA performance gains. This indirect evaluation is more diagnostic of real-world utility than standard pixel-level generation metrics (\eg, FVD, FID).
With such a joint evaluation of VLMs and WMs, a comprehensive assessment of Foresight Intelligence is conducted, aiming to answer two fundamental questions:
(1) What is the capability of VLMs in Foresight Intelligence?
(2) Can the data generated by WMs contribute to Foresight Intelligence?

Our contribution can be summarized as:
\begin{itemize}
    \item We propose \dataset, a dataset designed to train and evaluate the Foresight Intelligence of VLMs through tasks requiring counterfactual simulation and multi-agent causal reasoning---capabilities that go  beyond the reactive decision-making scope of existing VQA benchmarks.  Our experiments further demonstrate that \dataset can effectively enhance VLMs' foresight reasoning.
    \item {We introduce a proxy WM-evaluation protocol that measures whether predicted
    future videos and trajectories provide useful semantic cues for downstream
    foresight QA, while acknowledging that this protocol is probe-model dependent.}
    \item A comprehensive evaluation across multiple VLMs and WMs on \bench reveals that current models consistently struggle with high-level counterfactual reasoning, while WM-generated futures provide measurable but architecture-dependent semantic gains.
\end{itemize}

\section{Related Work}

\subsection{Spatio-Temporal Reasoning in VLMs}
Modern VLMs have progressed significantly beyond single-frame understanding~\cite{wang2022internvideogeneralvideofoundation,zhang-etal-2023-video}. Advanced models~\cite{maaz-etal-2024-video} now demonstrate a strong capacity to capture both temporal information and static spatial relations in videos, achieving robust performance in open dialogue and multimodal reasoning. For instance, Qwen2.5-VL~\cite{bai2025qwen25vltechnicalreport} leverage timestamp encoding for long video understanding, while InternVL-3.5~\cite{wang2025internvl35advancingopensourcemultimodal} employs a visual resolution router to efficiently process fine-grained video details. Notably, recent work such as GPT-Driver~\cite{mao2023gpt}, LanguageMPC~\cite{mao2023language}, EMMA~\cite{zhao2024equivariant}, and SimLingo~\cite{renz2025simlingo} has demonstrated that VLMs can perform chain-of-thought reasoning for autonomous driving, yet these models are primarily evaluated on present-state 
understanding rather than anticipating unobserved future events. 
Beyond VLM-based driving reasoning, trajectory forecasting methods model future motion using interaction-aware and spatio-temporal consistency cues~\cite{li2025future,li2021spatial}. 
However, these methods primarily output geometric trajectories or short-term motion states, whereas \FI requires language-grounded semantic reasoning about long-horizon, multi-agent, and counterfactual future consequences.

\subsection{Benchmarks}
To evaluate VLMs' capabilities, a series of benchmarks have been proposed, but these largely focus on understanding the present state.
\begin{itemize}
    \item \textbf{Static Spatial Reasoning:} Existing benchmarks range from 2D relational 
reasoning datasets such as CLEVR~\cite{8099698} and GQA~\cite{8953451} to static 3D 
spatial reasoning benchmarks such as ScanQA~\cite{9878756}, VSI-Bench~\cite{yang_thinking_2025}, 
and MMSI-Bench~\cite{yang2025mmsibenchbenchmarkmultiimagespatial}. 
Recent monocular and open-vocabulary works~\cite{wang2025monosr,li2025seeground,cheng2026perception} 
further advance spatial reasoning and grounding from visual observations, but primarily 
target current-scene understanding rather than long-horizon foresight.

    \item \textbf{Dynamic Spatial Reasoning:} Acknowledging the limitations of 
    static scenes, newer benchmarks have begun to address dynamics~\cite{9577425,
    zhang_dsi-bench_2025}. VLM4D~\cite{zhou2025vlm4dspatiotemporalawarenessvision} 
    treats video as a four-dimensional modality to probe spatiotemporal correlations. 
    However, this shift to dynamic scenes has exposed a critical flaw: multimodal 
    hallucination~\cite{ma20253dsrbenchcomprehensive3dspatial,wang2025orient,
    li-etal-2023-evaluating} has demonstrated that VLMs are highly prone to 
    orientation hallucinations and semantic priors, especially when processing 
    dynamic scenarios or uncommon viewpoints.

    \item \textbf{Driving VQA:} NuScenes-QA~\cite{qian2024nuscenes} targets 
    static perception of the current frame. DriveLM~\cite{sima2024drivelm} chains 
    perception and planning, but restricts queries to immediate control actions 
    within short horizons. OmniDrive~\cite{wang_omnidrive_2025} further introduces 
    counterfactual queries, yet outputs geometric waypoints evaluated via L2 
    displacement error---grounding evaluation in path generation rather than 
    causal scene understanding. \dataset departs from this paradigm: it targets 
    natural-language causal judgments over multi-agent counterfactual scenarios, 
    a semantic simulation-level capability not addressed by prior nuScenes-based 
    benchmarks (see \cref{tab:superiority_comparison}).
\end{itemize}

\subsection{Future-Oriented World Models}

A core application of WMs is the generation of physically plausible future videos for autonomous driving. Early approaches based on diffusion models~\cite{9878449,10655388} suffered 
from limited controllability and short rollout horizons; subsequent GPT-style 
autoregressive frameworks~\cite{yan2021videogptvideogenerationusing,10297415,
chen2025deepverse4dautoregressivevideo} improved temporal consistency and 
enabled extended multi-frame generation under continuous control.

Among the latest advancements, Epona~\cite{Zhang_2025_ICCV} and DrivingWorld~\cite{hu_drivingworld_2024} represent two representative directions: Epona focuses on large-scale video-to-action pretraining for long-horizon driving imagination, while DrivingWorld incorporates explicit spatial constraints and controllable rollout strategies to stabilize world model predictions.
Despite these advancements, the emphasis of current approaches remains centered on prediction. Although the synthesized videos often appear visually realistic, they frequently lack semantic fidelity, raising concerns about whether the predicted futures align with actual causal dynamics and scene semantics~\cite{bruce2024genie}.

In this work, we introduce \bench, a unified benchmark designed to address the above limitations and enable a thorough assessment of the foresight capabilities of both VLMs and WMs.

\section{\dataset}

\begin{table}[t]
\caption{\textbf{Comprehensive comparison between \dataset and existing driving VQA benchmarks.} \dataset uniquely targets long-horizon foresight intelligence, enabling 
joint evaluation of both VLMs and World Models' semantic consistency.}
\label{tab:superiority_comparison}
\centering
\resizebox{\textwidth}{!}{
\begin{tabular}{l | c c c c c c}
\toprule
\textbf{Feature / Benchmark} & \textbf{NuScenes-QA} & \textbf{DriveLM} & \textbf{OmniDrive} & \textbf{DriveGPT4} & \cellcolor{navyblue!5}\textbf{\dataset (Ours)} \\
\midrule
Primary Focus & Perception & Planning & Planning & Action & \cellcolor{navyblue!5}\textbf{Foresight} \\
Long-horizon Reasoning (\textbf{12s}) & $\times$ & $\times$ & $\times$ & $\times$ & \cellcolor{navyblue!5}\checkmark \\
Counterfactual (What-if) Reasoning & $\times$ & $\times$ & Partial & $\times$ & \cellcolor{navyblue!5}\checkmark \\
Hierarchical Cognitive Tasks (\textbf{9 Levels}) & $\times$ & $\times$ & $\times$ & $\times$ & \cellcolor{navyblue!5}\checkmark \\
{Multi-Agent-Aware Counterfactual QA} & $\times$ & Partial & \checkmark & $\times$ & \cellcolor{navyblue!5}\checkmark \\

\textbf{Foresight-oriented Training Split} & $\times$ & $\times$ & $\times$ & $\times$ & \cellcolor{navyblue!5}\checkmark \\
\textbf{World Model Semantic Eval} & $\times$ & $\times$ & $\times$ & $\times$ & \cellcolor{navyblue!5}\checkmark \\
Dataset Source & nuScenes & nuScenes & nuScenes & nuScenes & \cellcolor{navyblue!5}nuScenes \\
\bottomrule
\end{tabular}
}
\end{table}

\subsection{Foresight Intelligence}
\label{sec:foresight intelligence}
We use the term \FI to encapsulate the advanced predictive capabilities that go beyond simple pattern recognition or short-term forecasting. While ``prediction'' is often used in machine learning for tasks like next-token generation or immediate trajectory forecasting, ``foresight'' implies a deeper, more causal understanding. It is the ability to model, simulate, and evaluate complex, long-horizon futures, often involving multiple interacting agents and counterfactual ``what-if'' scenarios. 

\FI is conceptually distinct from the planning-oriented reasoning studied 
in prior datasets.
Planning-oriented benchmarks and agentic systems, such as 
DriveLM~\cite{sima2024drivelm}, OmniDrive~\cite{wang_omnidrive_2025}, 
and One Agent~\cite{li2026one}, mainly focus on immediate decisions 
or short-horizon behaviors under observable contexts. 
\FI, by contrast, requires anticipating long-horizon semantic consequences 
under multi-agent uncertainty and counterfactual conditions.
This constitutes a qualitatively distinct capability: \emph{simulation-level cognition} rather than reaction-level control.%

Motivated by insights from cognitive neuroscience \cite{schacter_remembering_2007} and the demands of autonomous decision-making \cite{712192}, we posit that \FI is supported by three foundational pillars:
\begin{itemize}
    \item Situational Modeling. The ability to build a structured, semantic understanding of the current world state, including agent relationships and environmental constraints.
    \item Causal and Dynamic Simulation. The core predictive engine that generates physically plausible and temporally coherent futures, including the ability to run ``what-if'' counterfactual scenarios based on causality.
    \item \textbf{Goal-Oriented Evaluation.} The capacity to reason over 
    simulated futures to assess risks and opportunities, enabling selection 
    of an optimal action that ensures safety and achieves a specific objective.
\end{itemize}
These three pillars collectively guide the design of \bench and are intrinsically woven into every task.

\subsection{Overview}
\label{sec:dataset_overview}
We introduce \dataset (see \cref{fig:teaser}), a VQA dataset designed to train and quantitatively evaluate Foresight Intelligence in autonomous driving scenarios, where vehicles must operate in highly dynamic environments and anticipate future situations to ensure safety.
\dataset aims to address two fundamental questions:
 (1) Whether current VLMs, which are primarily evaluated on present-state understanding, possess the capability for Foresight Intelligence; and
 (2) Whether the visually realistic data generated by WMs can effectively contribute to developing Foresight Intelligence.

To address these two questions, we design two specialized evaluation paradigms on \bench. Both are cast as a question--answering task in which the model must answer future-related queries based on historical visual observations and motion trajectories.

\textbf{VLM-oriented evaluation} This evaluation is designed to assess the foresight understanding capability of a pretrained VLM ($\mathbf{\mathcal{VLM}}$):
\begin{equation}
    \hat{a} = \mathbf{\mathcal{VLM}}(q, V_{-T_{h}:0}, Traj_{-T_{h}:0})
    \label{eq:vlm_eval}
\end{equation}
where $\hat{a}$ is the predicted answer, $q$ denotes the textual query, $V_{-T_h:0}$ refers to the historical video observations from $t=-T_h$ to the current time step ($t=0$), and $Traj_{-T_h:0}$ represents the associated motion trajectories.

\textbf{WM-oriented evaluation.} This evaluation assesses the semantic coherence of WM-generated futures via a proxy-evaluation paradigm: we measure the downstream QA performance gains of VLMs when augmented with WM outputs, treating this gain as a proxy for the semantic utility of the predicted content.
\begin{equation}
    \hat{a} = \mathbf{\mathcal{VLM}}(q, V_{-T_{h}:0}, Traj_{-T_{h}:0}, \hat{V}_{1:T_f}, \hat{Traj}_{1:T_f})
    \label{eq:wm_eval}
\end{equation}
\begin{equation}
   \hat{V}_{1:T_f}, \hat{Traj}_{1:T_f} = \mathcal{WM}(V_{-T_{h}:0}, Traj_{-T_h:0})
\end{equation}
where $\hat{V}_{1:T_f}$ and $\hat{Traj}_{1:T_f}$ are predicted future video and trajectory from WM ($\mathcal{WM}$).

\begin{figure}[t]
  \centering
   \includegraphics[width=0.95\linewidth]{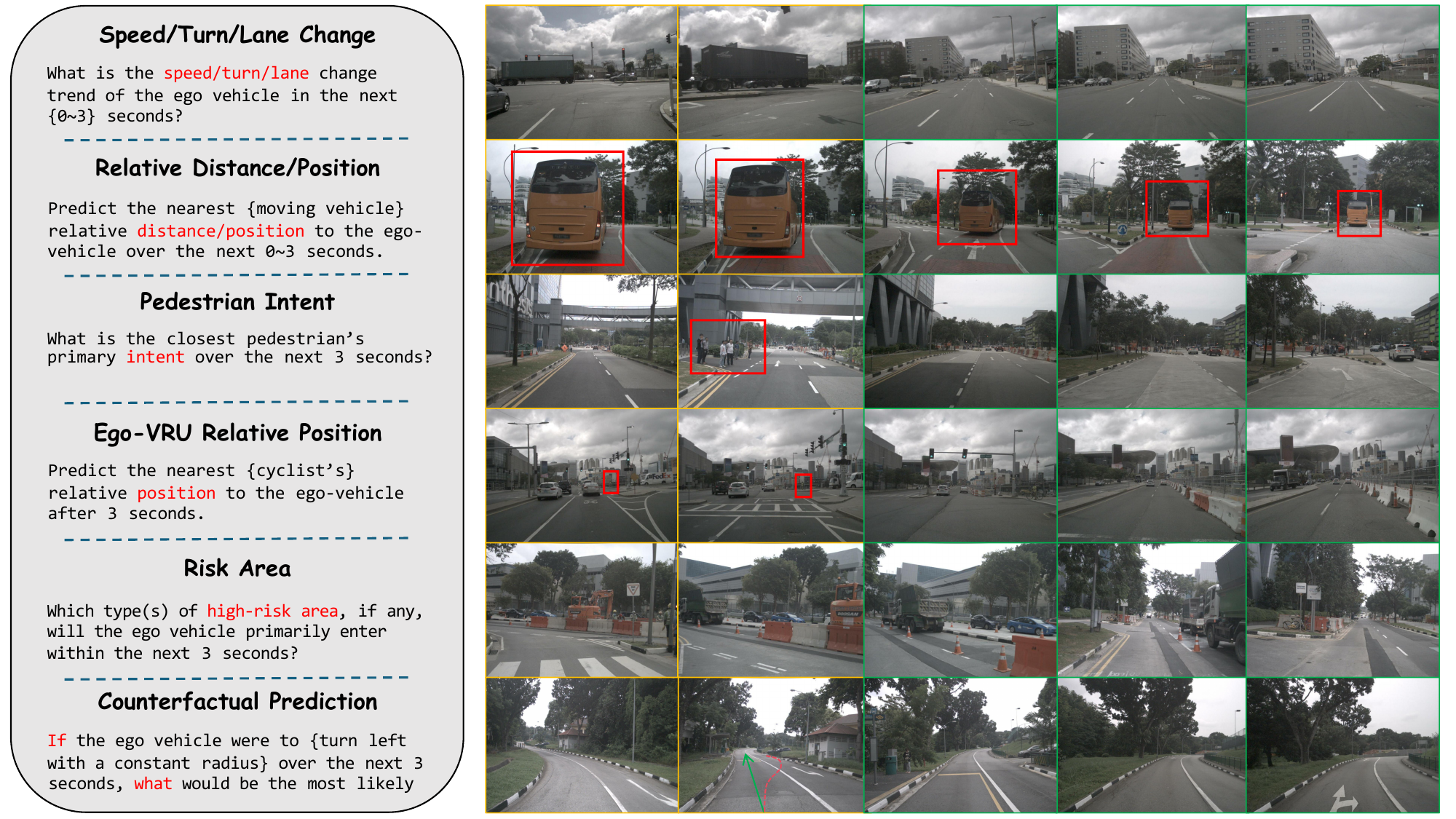}
   \caption{\textbf{Visualization of \bench.} Yellow boxes represent historical frames, and green boxes represent future frames. The bounding boxes are illustrative only and do not appear in the actual dataset. Content in \{\} is filled into a sentence template. Note: the questions above are simplified slightly for clarity and brevity.}
   \label{fig:Visualization}
\end{figure}

\section{Evaluation on \bench}

\subsection{Dataset Curation \& Annotations}
\begin{figure}[!t]
  \centering
   \includegraphics[width=0.95\linewidth]{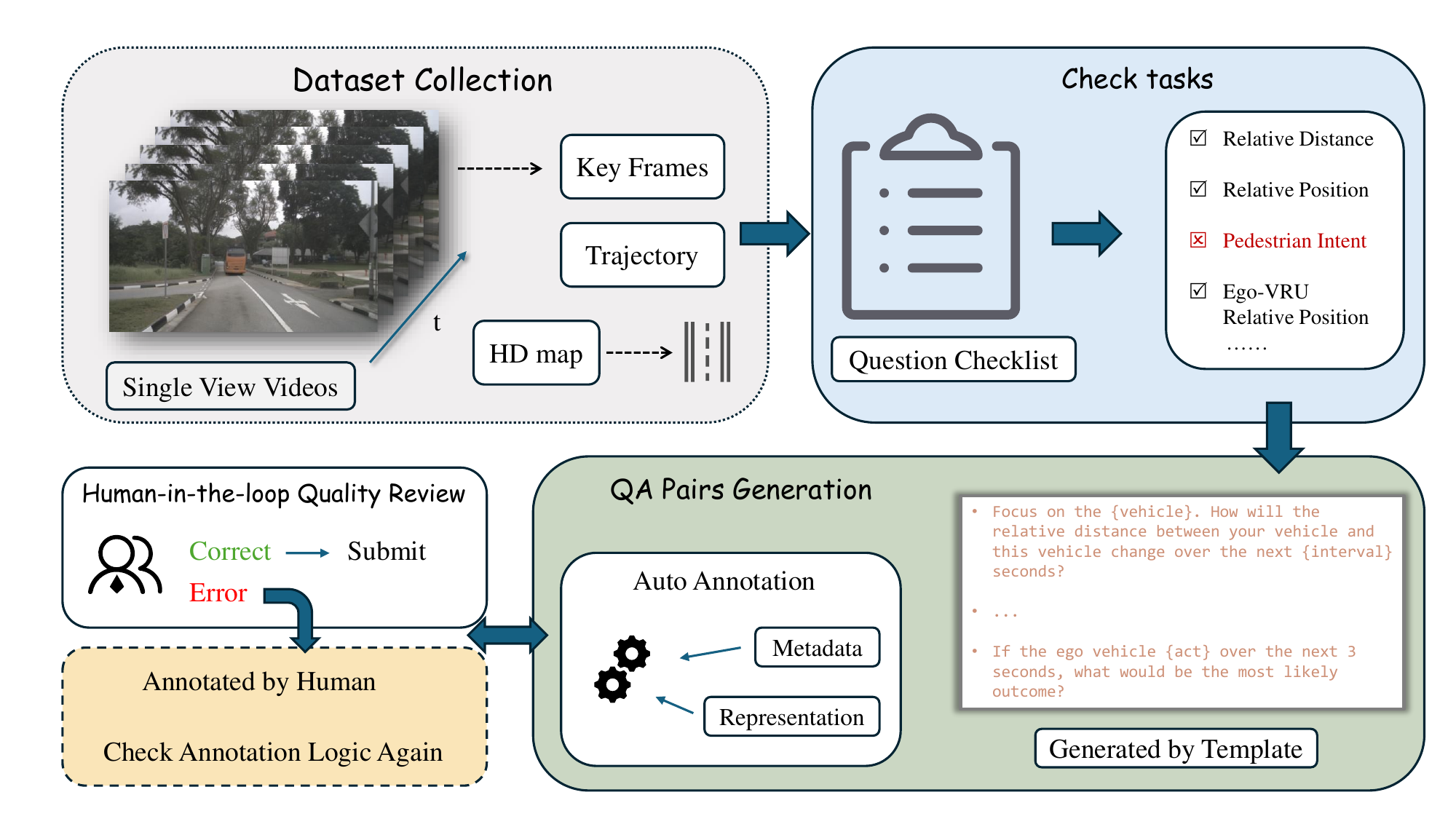}
   \caption{\textbf{\dataset construction pipeline.} Note: the questions above are simplified slightly for clarity and brevity.}
   \label{fig:construction pipeline}
\end{figure}

Building upon the nuScenes dataset \cite{caesar_nuscenes_2020}, which provides rich sensory and map information, we develop an automatic data annotation pipeline to generate a high-quality dataset enabling both training and systematic evaluation of Foresight Intelligence, as illustrated in \cref{fig:construction pipeline}.

\textbf{Scene Pre-processing and Representation.} To approximate the driver's perspective, we use videos recorded from the front-facing camera, each approximately 15 seconds long. We segment each video into a 3-second historical observation window and a 12-second future window. Because not every scene is valid for all tasks, we apply filtering to ensure annotation quality, yielding 18 to 28 questions per scene. The HD map is represented using lane lines.

\textbf{Cognition-Inspired Task Design.} To comprehensively train and evaluate Foresight Intelligence, we design nine tasks within \dataset (see \cref{fig:Visualization}), grouped into three major categories that span from low-level perception to high-level reasoning:

\textit{Spatio-temporal Dynamic Reasoning (low-level).}This subset assesses the model's capacity to capture dynamic spatio-temporal relations over long time horizons. While individual sub-tasks such as speed and lane-change prediction are studied in motion forecasting literature, our key distinction is the \emph{QA formulation}: models must produce interpretable language-grounded answers about ego-vehicle motion states and agent interactions over a 12-second horizon, without access to ground-truth future observations. This formulation bridges motion forecasting with semantic scene understanding, enabling diagnosis of foresight-oriented reasoning capabilities.
It involves predicting the ego vehicle's future motion states (both longitudinal and lateral) and understanding its interactions with nearby moving agents, characterized by variations in relative distance and position. This level comprises five tasks: Speed Change, Turn Change, Lane Change, Relative Distance (Rel.~Dist.), and Relative Position (Rel.~Pos.).

\textit{VRU-centric Risk Assessment (mid-level).}
This level assesses the model's capability to detect and interpret Vulnerable Road Users (VRUs) and identify potential risk zones. It consists of three core annotation tasks:
\begin{itemize}
    \item Pedestrian Intent (Ped.~Int.).
Annotate the intent of the closest front-view pedestrian within the next 0--3 seconds, based on trajectory and map information.
    \item Ego--VRU Relative Position Change (E--V Rel.~Pos.). Label the final relative position of the closest VRU with respect to the ego vehicle at the end of the 0--3 second horizon.
    \item Risk Area.
Classify the environmental risk level in the upcoming 3-second interval.
\end{itemize}

\textit{High-level Causal Reasoning (high-level).}
To mitigate nuScenes' inherent ``safety bias'', the limited presence of unsafe or hazardous driving cases, we introduce a Counterfactual Prediction (CFP) setting that evaluates a model's sensitivity to unsafe behaviors and its capacity for causal understanding. For instance, consider the query: ``If the ego vehicle were to turn left with a constant radius in the next 3 seconds, what would be the most likely outcome?'' Such prompts allow us to directly 
{assess the model's rule-grounded counterfactual outcome reasoning ability,}

whether it can anticipate the safety implications of a prescribed action, such as predicting if a collision would occur.

\textbf{Counterfactual Annotation Validity.}
Generating reliable ground truth for counterfactual scenarios (i.e., what \emph{would} happen under a prescribed but unexecuted action) is inherently challenging. We address this through two complementary mechanisms.
First, our rule-based criteria are grounded in the closed-world geometry of nuScenes: given accurate 3D object trajectories, HD-map lane boundaries, and kinematic states, collision outcomes under a prescribed maneuver (\eg, constant-radius left turn) can be determined with high geometric fidelity via spatial intersection checks—without requiring an external simulator.
Second, to empirically validate annotation quality, we conducted a human-in-the-loop verification study in which domain experts independently adjudicated a randomly sampled subset of 150 CFP annotations covering diverse scene types, achieving an agreement rate of 89\% with the automated labels (Cohen's $\kappa = 78\% $).
These results confirm that rule-based CFP annotation is reliable within the closed geometric constraints of the nuScenes environment.%

\textbf{QA Data Generation.} For all nine tasks, we design a comprehensive question checklist using multi-sensor data, 3D object annotations, HD-map representations, and trajectory information. This checklist is applied to each scene to determine the tasks for which annotations are feasible, after which questions are generated via predefined templates.
To produce the answers, we establish task-specific label systems and rule-based criteria. We extract and analyze ground-truth information, including kinematic states, spatio-temporal relations, VRU behaviors, environmental risks, and counterfactual verification results, from the dataset and HD-map. Mapping this information to the defined criteria enables systematic derivation of accurate answers for every question.

\begin{wrapfigure}{r}{0.45\textwidth} 
  \centering
  \vspace{-15pt} 
  \includegraphics[width=\linewidth]{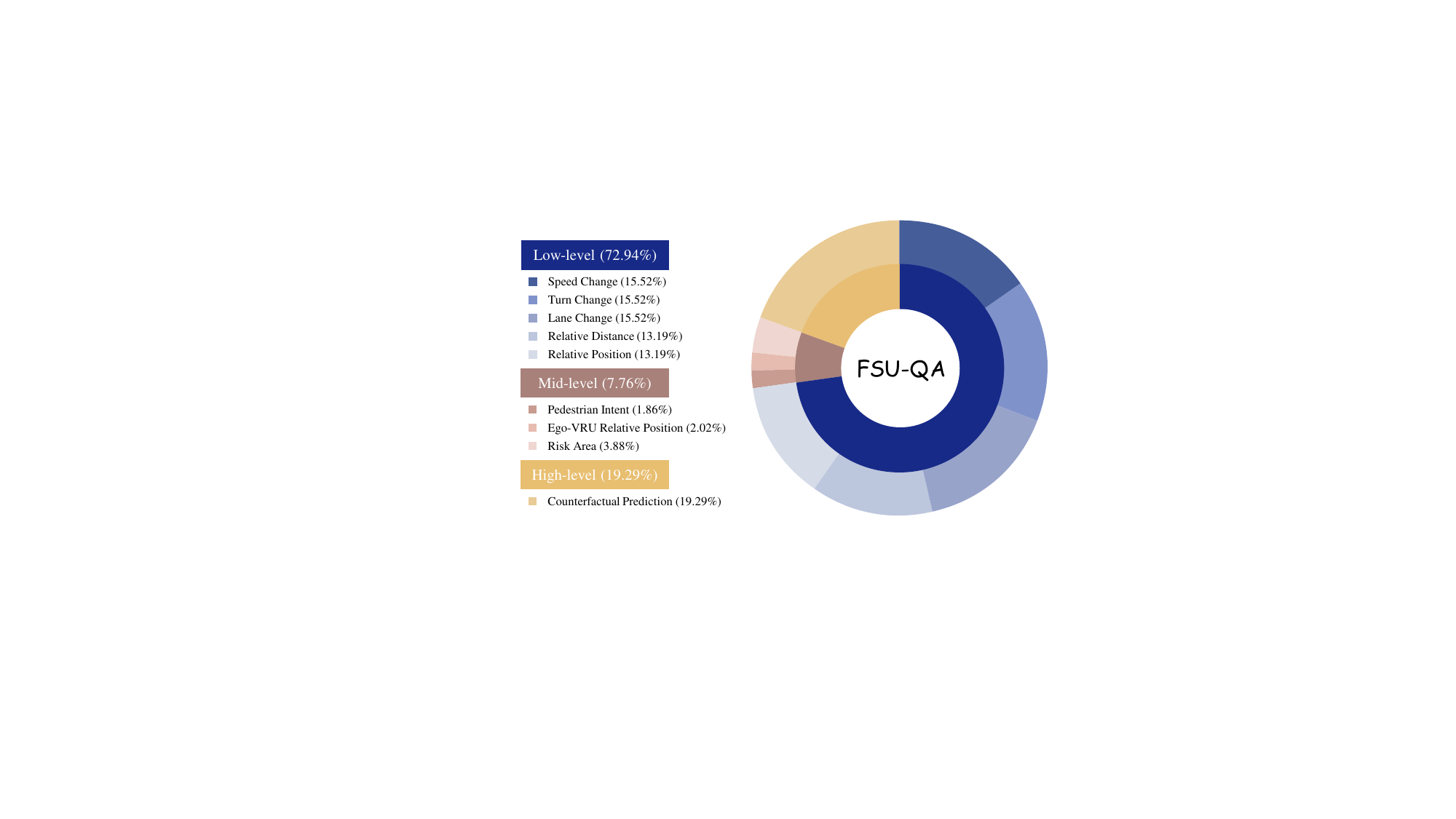}
  \caption{\textbf{\dataset Statistics.} The distribution of tasks across three levels.}
  \label{fig:dataset_statistics}
  \vspace{-10pt} 
\end{wrapfigure}

Specifically, we partition the entire video sequence and trajectory data into $N$ segments of 3 seconds each, denoted as $S_{1:N}$. The first segment $S_1$ is regarded as the historical observation, whereas the subsequent segments $S_{2:N}$ represent future intervals. For QA generation, we randomly select a timestep $i$ and construct a QA pair conditioned on $S_{1:i}$. This procedure yields QA pairs spanning varying temporal horizons, enabling evaluation of long-term foresight intelligence.

\textbf{Quality Control.}
Our quality control follows a structured human-in-the-loop verification framework. We randomly sampled 45 scenes (approximately 30\% of \bench), and three domain experts with backgrounds in autonomous driving independently reviewed each QA pair, categorizing flagged issues into: (i) label logic errors, (ii) edge-case misclassification, and (iii) question ambiguity. For each flagged category, we either corrected the annotation rule globally or handled individual corner cases manually. This two-way iteration was run for three rounds until the human-flagged error rate dropped below 5\%, at which point the evaluation set was considered finalized. In total, 654 QA pairs were revised or removed as a result of this process.

Additional information about the dataset curation process is provided in the \textit{Supplemental Materials}.

\begin{figure}[t!]
    \captionsetup{type=table}
    \centering
    \resizebox{0.95\textwidth}{!}{
    \begin{tabular}{r|cc|ccccccccc}
    & & &
    \rotatebox{75}{Speed Change} &
    \rotatebox{75}{Turn Change} &
    \rotatebox{75}{Lane Change} &
    \rotatebox{75}{Rel. Dist.} &
    \rotatebox{75}{Rel. Pos.} &
    \rotatebox{75}{Ped. Int.} &
    \rotatebox{75}{E-V Rel. Pos.} &
    \rotatebox{75}{Risk Area} &
    \rotatebox{75}{CFP} \\
    Methods & Rank & Overall & \multicolumn{5}{c}{\cellcolor{red!10}Low-level} & \multicolumn{3}{c}{\cellcolor{orange!10}Mid-level} & \multicolumn{1}{c}
    {\cellcolor{yellow!10}High-level}\\
    \hline
    \rowcolor{navyblue!5}
    \multicolumn{1}{l|}{\textcolor{black}{\textit{Open-source Models}}} & & & & & & & & & & & \\
    Qwen2.5-VL-72B & \cellcolor{oai-green-200}{3} & 45.86 & 37.00 & \cellcolor{oai-gray-600}{90.17} & \cellcolor{oai-gray-600}{97.50} & 28.66 & 14.43 & 21.74 & 9.88 & \cellcolor{oai-gray-600}{64.67} & 10.31 \\
    Qwen2.5-VL-7B  &6&44.74&34.00&88.33&97.50&27.85&14.43& \cellcolor{oai-gray-600}{37.68}&3.70&63.33&8.43 \\
    Qwen3-VL-32B & 6 & 44.74 & 33.83 & 83.00 & 95.17 & 38.01 & 14.43 & 17.39 & 6.17 & 56.00 & 11.11 \\
    Qwen3-VL-8B &12 & 41.48 &31.67&76.50&97.00&32.93&14.43&20.29&3.70&45.33&5.35\\

    Llama 4 Maverick & 4&45.47& \cellcolor{oai-gray-600}{41.83}&66.00&92.00&33.74&18.50&36.23&18.52&42.67& \cellcolor{oai-gray-600}{24.36}\\
    Llama 4 Scout & 11&41.95&41.67&70.33&81.50&30.69&14.43&36.23&3.70&50.67&16.06\\
    \hline
    \rowcolor{navyblue!5}
    \multicolumn{1}{l|}{\textcolor{black}{\textit{Closed-source Models}}} & & & & & & & & & & & \\
    GPT-4o Mini & 8 & 44.22 & 39.17 & 86.33 & 93.33 & 33.94 & 14.02 & 18.84 & 3.70 & 36.67 & 13.65 \\
    GPT-5 & \cellcolor{oai-green-600}{1} & 48.66 & 34.67 & 75.50 & 93.00 & \cellcolor{oai-gray-600}{39.63} & \cellcolor{oai-gray-600}{32.32} & 36.23 & \cellcolor{oai-gray-600}{46.91} &48.67 &20.75 \\
    Gemini-2.0 Flash & 13 & 41.43 & 32.00 & 74.17 & 91.17 & 33.54 & 15.04 & 30.43 & 7.41 & 29.33 & 12.45 \\
    Gemini-2.5 Flash & 9 & 44.04 & 38.83 & 67.33 & 94.83 & 30.69 & 20.73 & 23.19 & 27.16 &54.67 &14.46 \\
    Gemini-2.5 Pro & 9 & 44.04 & 37.83 & 66.17 & 91.67 & 35.37 & 22.76 & 30.34 & 37.04 & 25.33 & 18.47 \\
    Claude-3.7-Sonnet& 5 & 45.00 & 31.83 & 84.67 & 92.67 & 35.16 & 16.67 & 33.33 & 19.75 & 44.00 & 14.59 \\
    Claude-Sonnet-4.5 & \cellcolor{oai-green-400}{2} & 46.38 & 37.17 & 76.67 & 94.50 & 35.77 & 20.33 & 27.54 & 14.81 & 49.33 & 19.54 \\
    \hline
    \rowcolor{navyblue!5}
    \multicolumn{1}{l|}{\textcolor{black}{\textit{Finetuned Model}}}& & & & & & & & & & & \\
    Qwen3-VL-8B-FI &-& 59.59&43.67 & 92.33& 93.17& 38.21& 39.63& 28.99&38.27 &66.00 &50.20 \\
    \end{tabular}
 }
    \caption{\textbf{Baseline Evaluation on \bench:} only input historical videos and trajectories. \colorbox{oai-gray-600}{Dark gray} cells indicate the best result across all models. }
    \label{table:baseline eval}
\end{figure}

\subsection{Statistics \& Analysis}

\dataset contains more than 21K question--answer pairs generated from 850 real-world driving videos in the nuScenes dataset \cite{caesar_nuscenes_2020}. This ensures that \dataset robustly covers a wide spectrum of real-world driving scenarios, including different times of day, adverse weather conditions, and diverse urban geographies (Boston and Singapore). Among them, 18K QA pairs (700 scenes) are allocated for training, and 3K+ QA pairs (150 scenes) are held out for evaluation (\bench). An overview of the \dataset statistics is provided in \cref{fig:dataset_statistics}.

\subsection{Evaluation Setup}
\textbf{Benchmark Models.} We conduct a comprehensive evaluation of a broad spectrum of Vision--Language Models (VLMs), spanning closed-source and open-source systems of varying parameter sizes and architectural types, to reveal their capabilities in foresight intelligence.

For closed-source models, we assess \textit{Gemini-2.0-Flash}, \textit{Gemini-2.5-Flash}, \textit{Gemini-2.5-Pro}~\cite{comanici2025gemini25pushingfrontier}, \textit{GPT-4o-Mini}~\cite{openai2024gpt4ocard}, \textit{GPT-5}, \textit{Claude-3.7-Sonnet-Standard} and \textit{Claude-Sonnet-4.5}.

For open-source models, we evaluate \textit{Qwen2.5-VL-72B-Instruct}, \textit{Qwen2.5-VL-7B-Instruct}~\cite{bai2025qwen25vltechnicalreport}, \textit{Qwen3-VL-32B-Instruct}, \textit{Qwen3-VL-8B-Instruct}, \textit{Llama-4-Scout}, and \textit{Llama-4-Maverick}~\cite{meta2025llama4}.

To assess models' capability to generate semantically coherent future data, we employ two representative world models: \textit{Epona}~\cite{Zhang_2025_ICCV}, which integrates autoregressive and DiT components, and \textit{DrivingWorld}~\cite{hu_drivingworld_2024}, a purely autoregressive architecture. In the experiments, all previously mentioned VLMs serve as proxy evaluators, enabling a comprehensive evaluation.

\textbf{Evaluation Protocol.}
To ensure reproducibility, we detail the full evaluation protocol here and in Appendix~A.
All VLMs are queried via their official APIs (or local inference for open-source models) using a zero-shot prompt format consisting of a task-specific system instruction, the encoded historical video frames, the trajectory representation, and the multiple-choice question.
Frames are sampled at 2 fps from the 3-second historical window, yielding 7 input frames per query.
Trajectories are encoded as a sequence of $(x, y, \theta)$ waypoints at 2 Hz.
For API-based models, including proprietary systems and hosted open-source
models, we use the default official API settings with a single-pass query
protocol and no retry policy. For locally deployed Qwen models, we use BF16
inference with temperature $=0.1$ in a single pass, which reduces repetitive
outputs observed under strict greedy decoding while keeping generation nearly
deterministic.
The full prompt template, including system instructions and input formatting, is provided in Appendix.

\textbf{Metric Design.} The questions are categorized into two types: Multiple-Select Questions (MSQ) and Single-Select Questions (SSQ). We adopt \textit{Accuracy} ($\mathcal{ACC}$) as the evaluation metric. An answer is considered correct only when it exactly matches the ground truth; partial matches are not counted as correct. For continuous-valued tasks (\eg, Relative Distance), choices are discretized into labeled bins; we report bin-level accuracy to maintain a unified QA evaluation framework.

\section{Main Results}

\begin{figure}[!t]
    \centering
    \setlength{\tabcolsep}{1pt}
    \begin{tabular}{cccc}
        \begin{subfigure}{0.25\textwidth}
            \includegraphics[width=\linewidth]{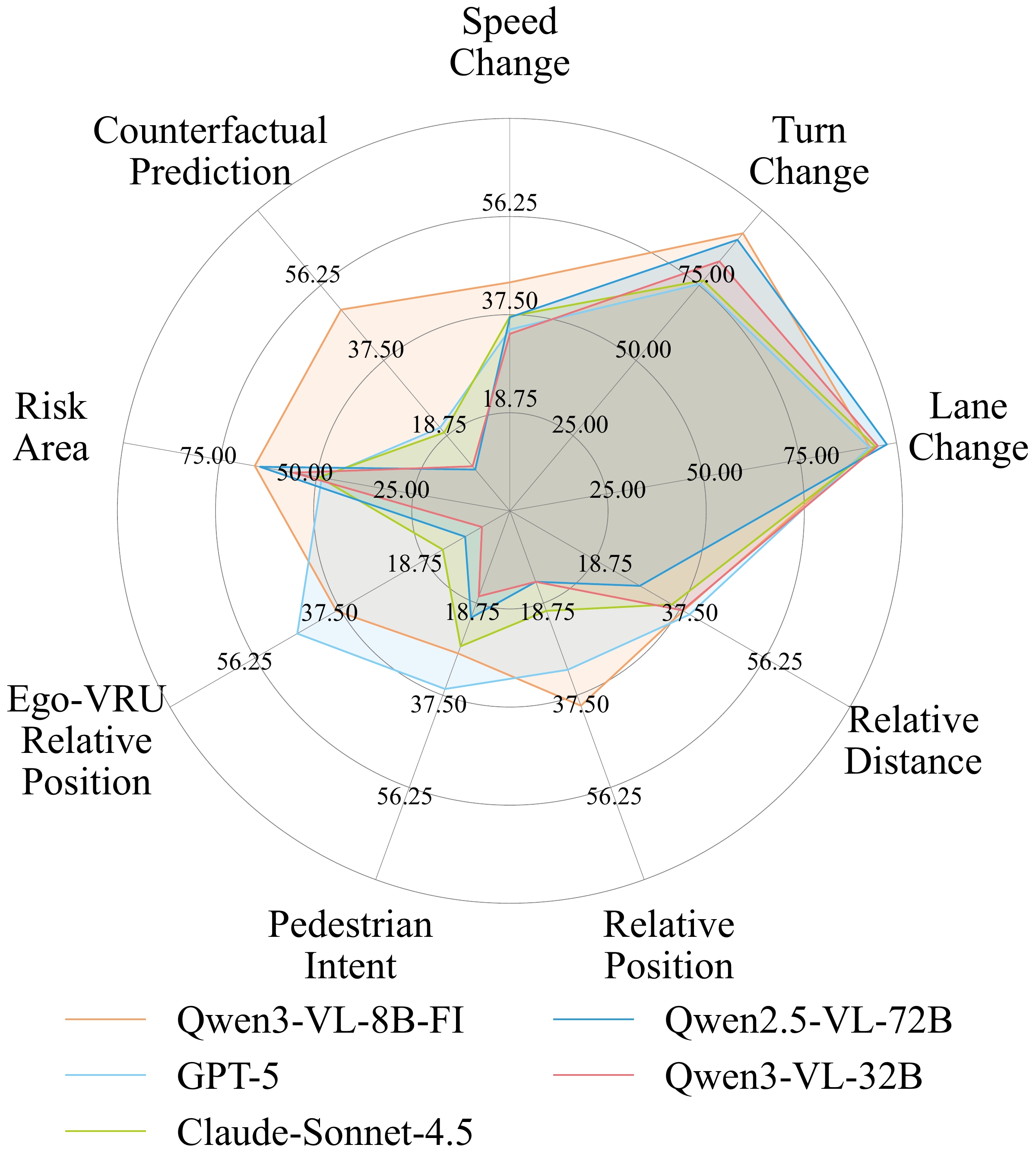}
            \caption{Baseline}
            \label{fig:epona_a}
        \end{subfigure} &
        \begin{subfigure}{0.25\textwidth}
            \includegraphics[width=\linewidth]{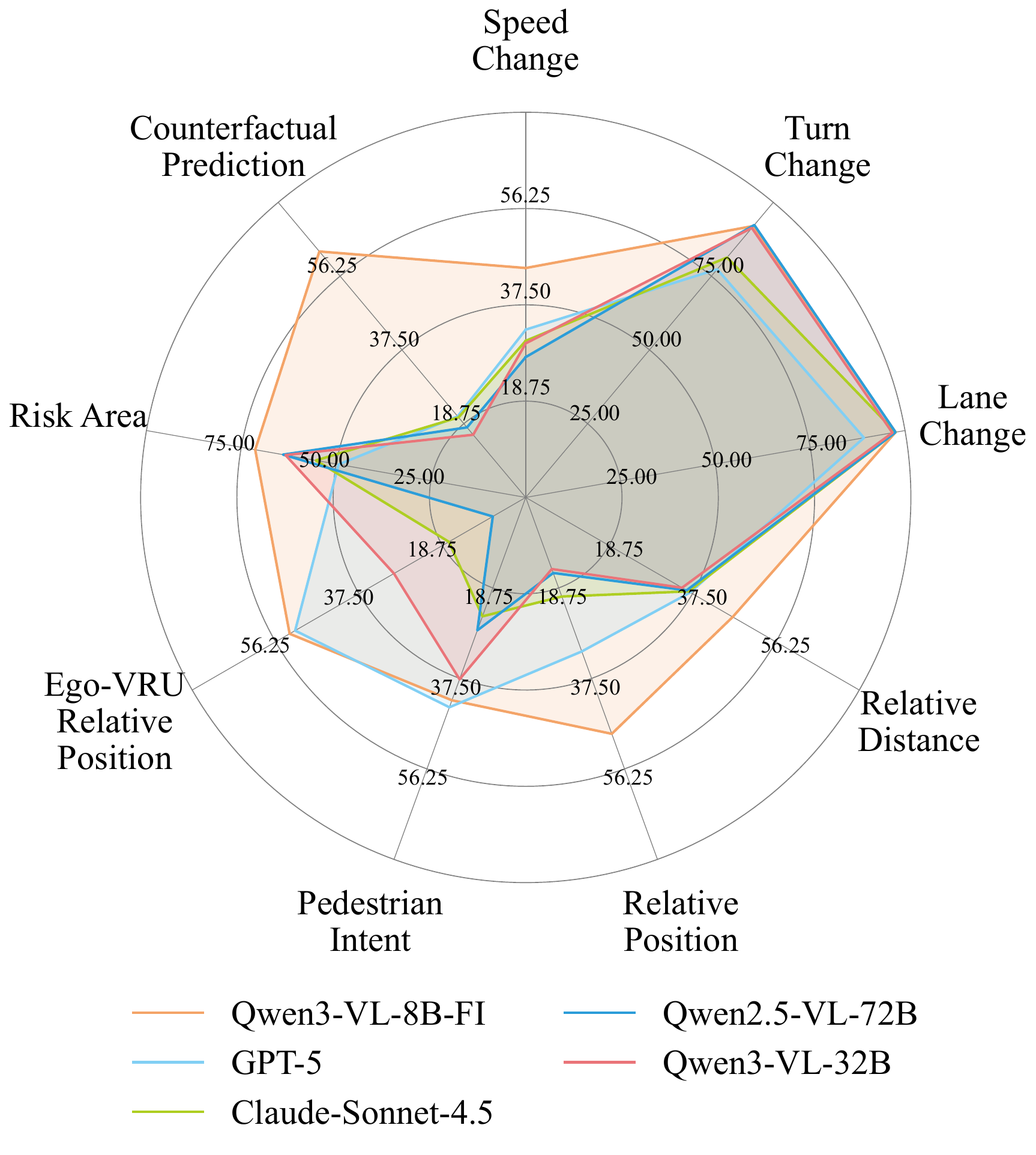}
            \caption{Base \& Pred. Video}
            \label{fig:epona_b}
        \end{subfigure} &
        \begin{subfigure}{0.25\textwidth}
            \includegraphics[width=\linewidth]{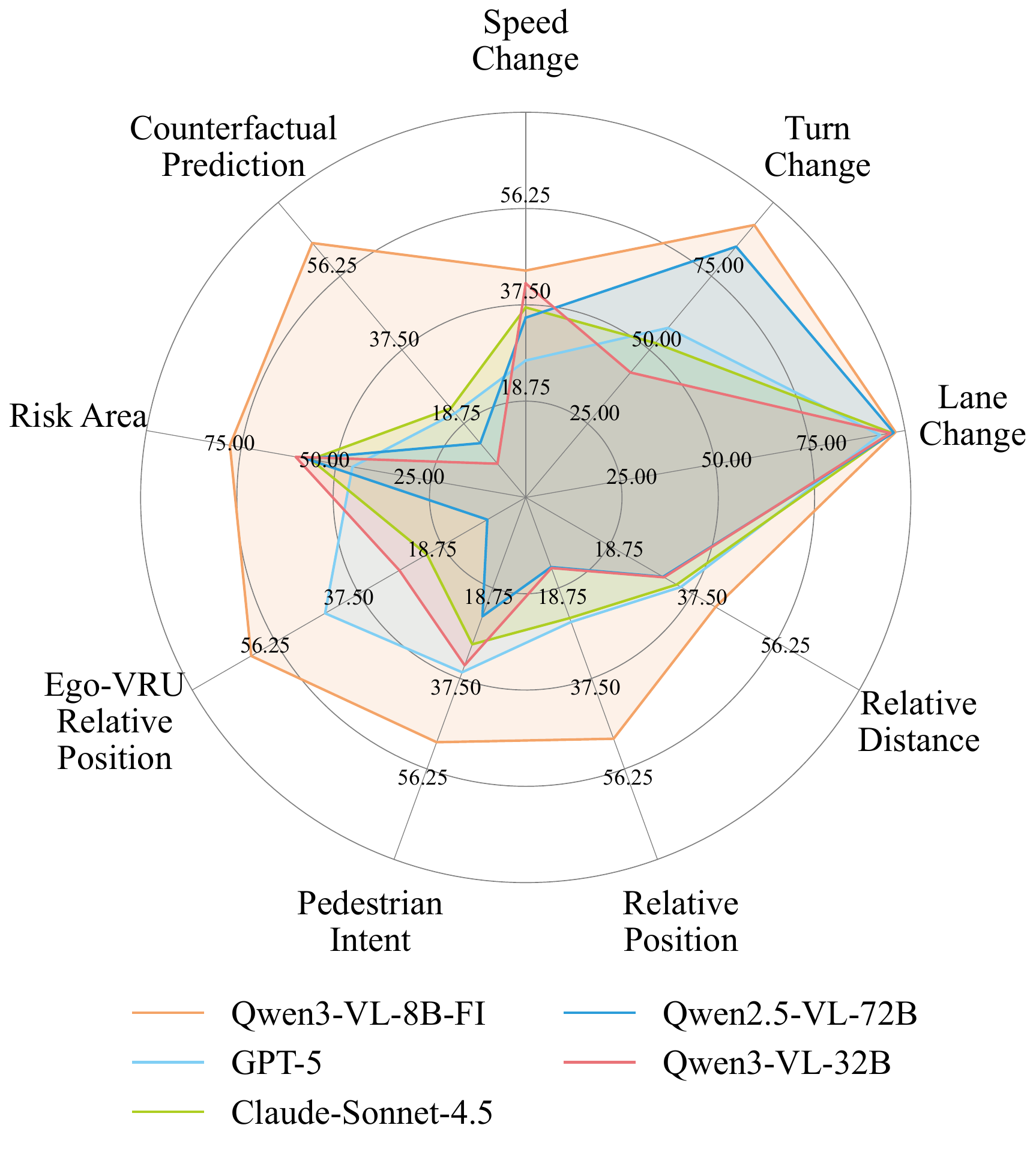}
            \caption{Base \& Pred. Traj}
            \label{fig:epona_c}
        \end{subfigure} &
        \begin{subfigure}{0.25\textwidth}
            \includegraphics[width=\linewidth]{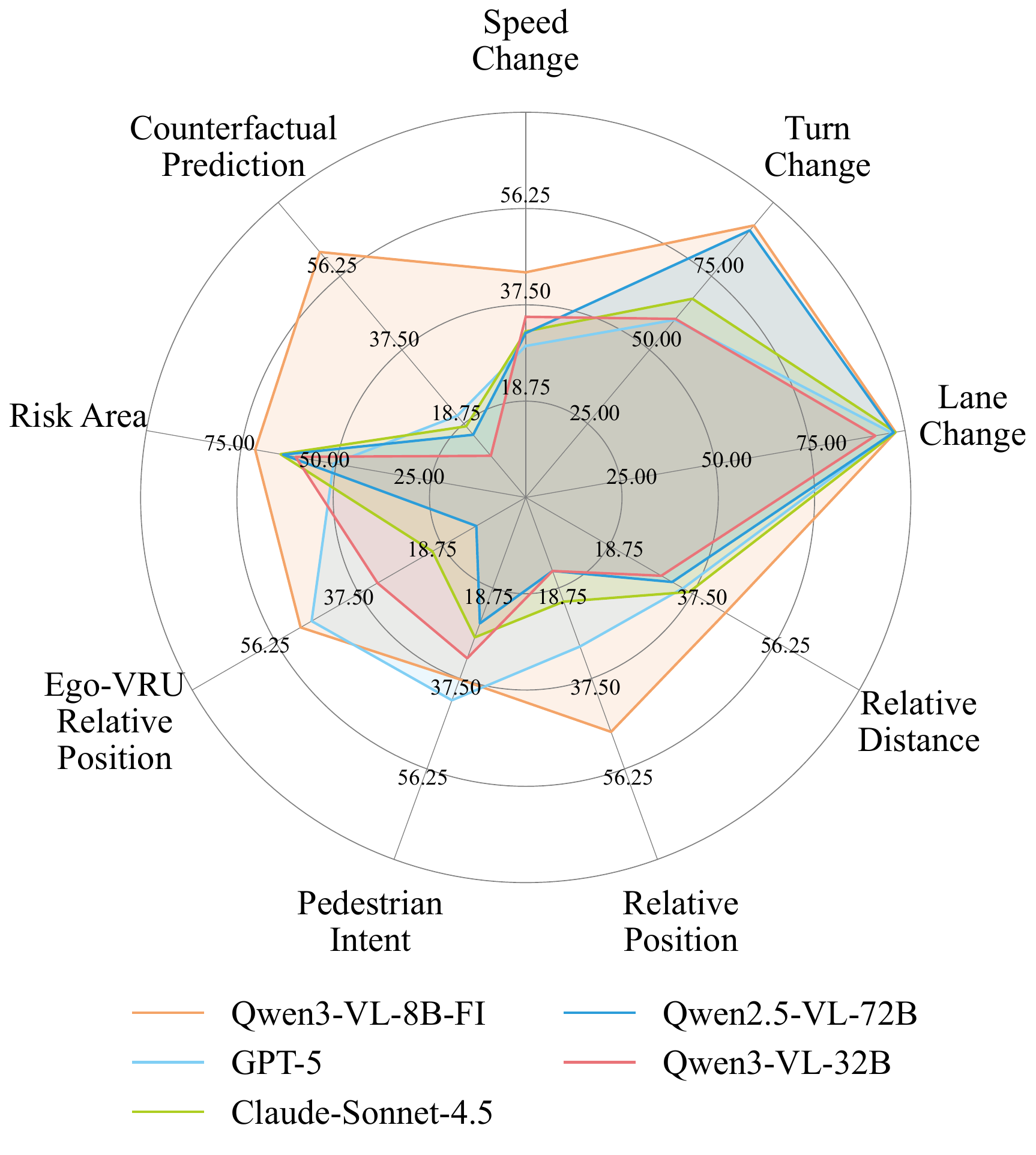}
            \caption{Base \& Pred. V+T}
            \label{fig:epona_d}
        \end{subfigure}
    \end{tabular}
    \caption{\textbf{Results with data generated by \textit{Epona}.} (a)--(d) correspond to results under four different inputs. }
    \label{fig:performance on epona}
\end{figure}

\begin{figure}[!t]
    \centering
    \setlength{\tabcolsep}{1pt}
    \begin{tabular}{cccc}
        \begin{subfigure}{0.25\textwidth}
            \includegraphics[width=\linewidth]{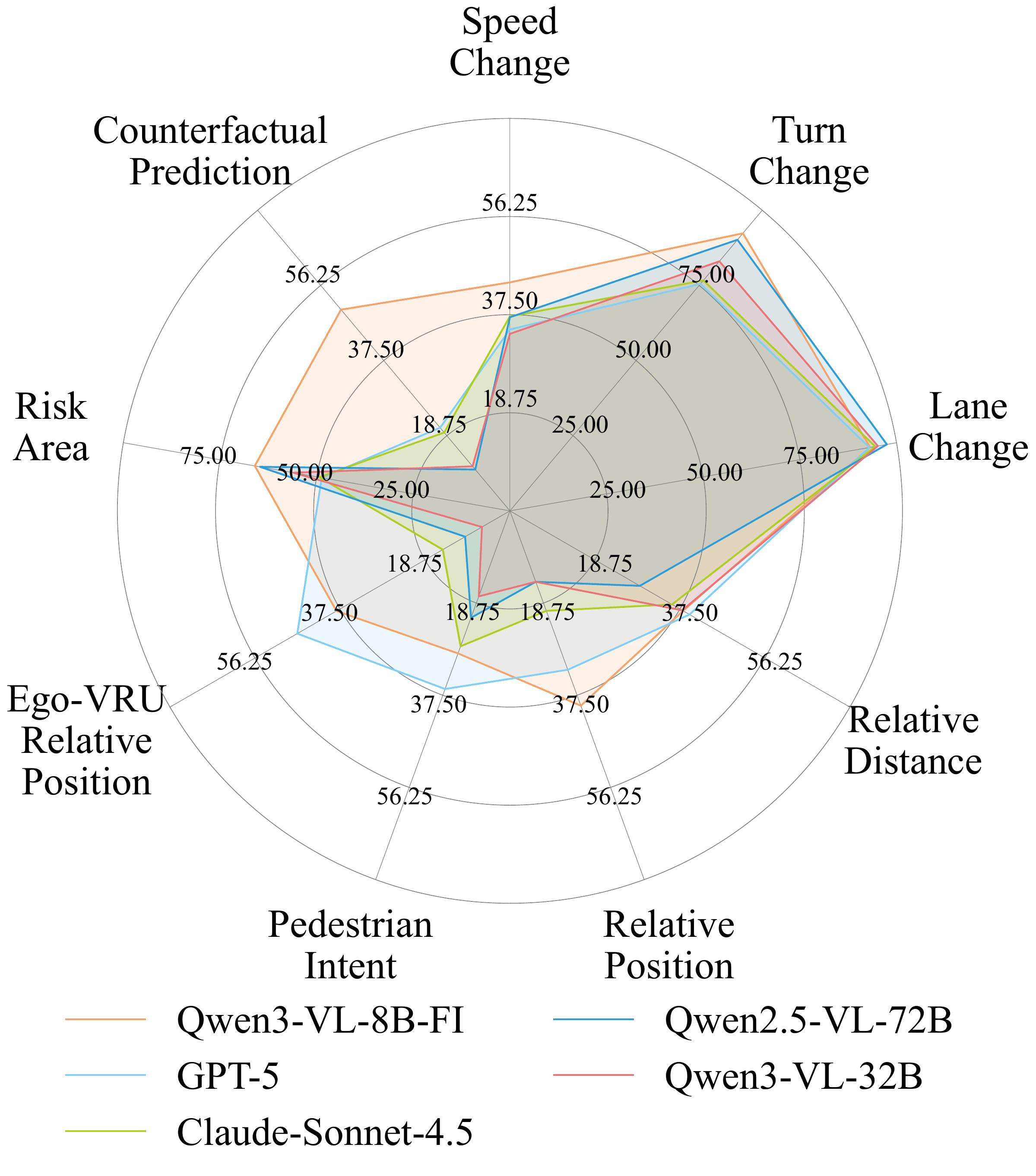}
            \caption{Baseline}
            \label{fig:dw_a}
        \end{subfigure} &
        \begin{subfigure}{0.25\textwidth}
            \includegraphics[width=\linewidth]{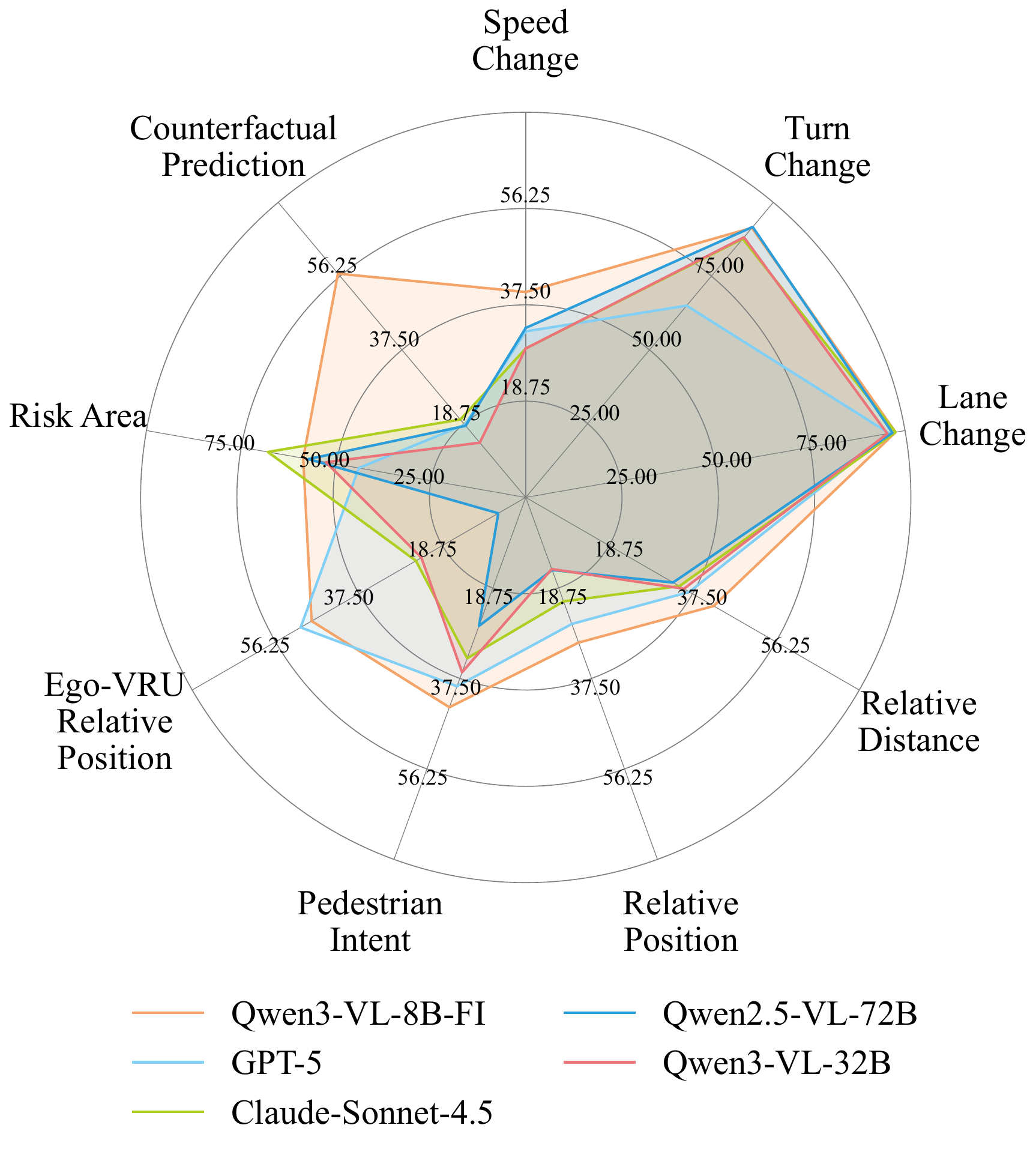}
            \caption{Base \& Pred. Video}
            \label{fig:dw_b}
        \end{subfigure} &
        \begin{subfigure}{0.25\textwidth}
            \includegraphics[width=\linewidth]{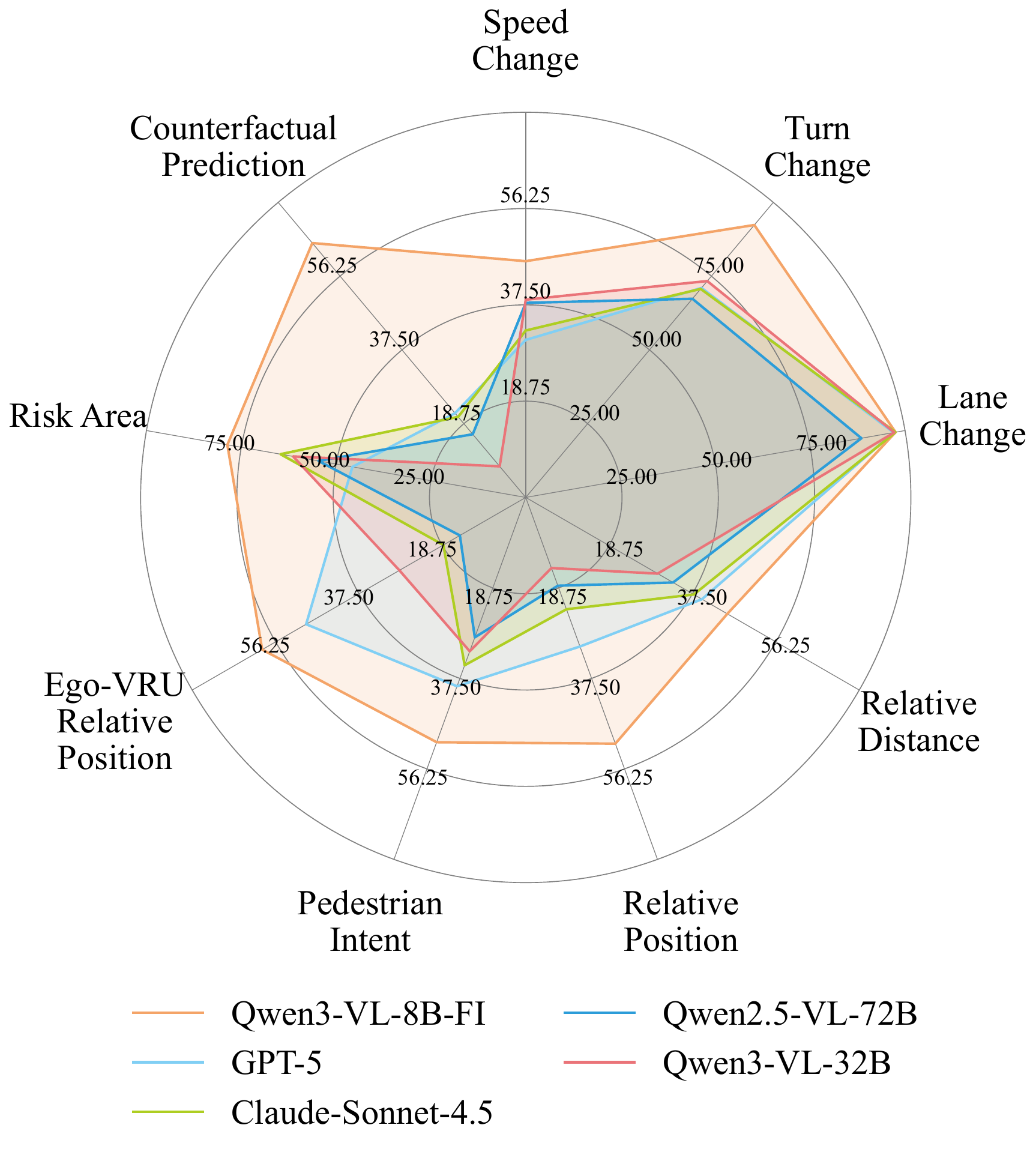}
            \caption{Base \& Pred. Traj}
            \label{fig:dw_c}
        \end{subfigure} &
        \begin{subfigure}{0.25\textwidth}
            \includegraphics[width=\linewidth]{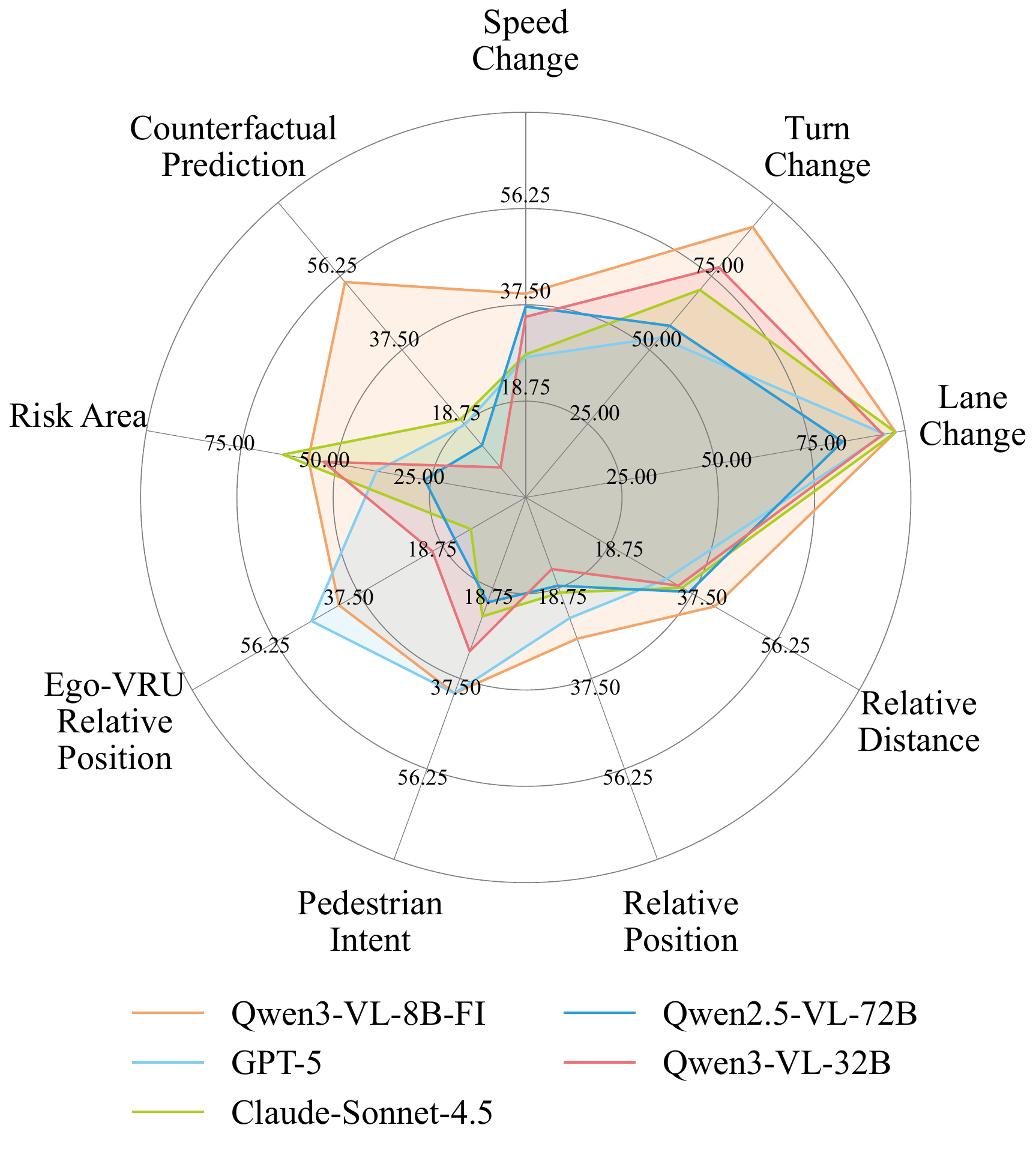}
            \caption{Base \& Pred. V+T}
            \label{fig:dw_d}
        \end{subfigure}
    \end{tabular}
    \caption{\textbf{Results with data generated by \textit{DrivingWorld}.} (a)--(d) correspond to results under four different inputs.}
    \label{fig:performance on drivingworld}
\end{figure}

We follow the joint evaluation pipeline described in Sec.~\ref{sec:dataset_overview}. Specifically, we first conduct the VLM-oriented evaluation to assess each VLM's foresight intelligence and establish a baseline for subsequent comparisons. We then perform the WM-oriented evaluation, where the semantic coherence of WM-generated future data is quantified by the performance gains over this baseline.

\subsection{VLM-oriented Evaluation}
As shown in Eq.~\ref{eq:vlm_eval}, the historical video sequence and trajectory are used as inputs to the VLM to generate future-related answers. 

\textbf{Closed-source Models.} Closed-source models demonstrate overall stronger foresight reasoning capabilities compared to most open-source counterparts, with GPT-5 achieving the highest overall score (48.66) and securing Rank 1 among all evaluated models. Notably, GPT-5 shows clear advantages in mid-level spatio-temporal reasoning, indicating superior ability to understand dynamic interactions between agents.

Claude-Sonnet-4.5 ranks 2nd overall, with a notable advantage on Risk Area (49.33)---the highest among closed-source models---suggesting heightened sensitivity to safety-critical scene cues. Meanwhile, Claude-3.7-Sonnet shows consistent mid-level performance but lags slightly behind Claude-Sonnet-4.5 in high-level reasoning.
In contrast, Gemini models (2.0 Flash, 2.5 Flash, 2.5 Pro) exhibit more conservative performance, with overall scores around 41--44. Gemini models score competitively on geometric tasks such as Lane Change (91--95) and Turn Change (66--74), but consistently underperform on Risk Area (25--55) and high-level CFP tasks (12--18), indicating limitations in safety-critical and counterfactual reasoning.
Finally, GPT-4o Mini, despite being a lightweight model, achieves competitive overall accuracy (44.22), surpassing some open-source models.
To better understand these performance differences, we analyze model-specific error patterns.
GPT-5 leads on both Rel. Pos. (32.32) and E-V Rel. Pos. (46.91), 
suggesting superior capacity for multi-agent interaction modeling 
and understanding.
These patterns suggest that foresight reasoning is not merely a 
function of model scale, but is sensitive to the richness of 
semantic scene understanding in the pretraining corpus.

\textbf{Open-source Models.} Among open-source models, Qwen2.5-VL-72B achieves the best overall performance (45.86), ranking 3rd across all models. It exhibits strong capabilities on Low-level foresight tasks, particularly in Turn Change (90.17) and Lane Change (97.50), indicating that large Qwen models excel at recognizing imminent maneuver intentions from short-term visual cues. However, its performance drops noticeably on mid-level tasks, suggesting limited understanding of multi-agent spatial dynamics. Medium-sized models such as Qwen2.5-VL-7B and Qwen3-VL-32B deliver competitive overall scores (44.74), showing that open-source models can remain effective even at smaller scales. Nevertheless, they generally exhibit weaker high-level reasoning, revealing challenges in long-term, safety-critical forecasting. The Llama 4 series shows different behavior: Llama 4 Maverick performs well on mid-level tasks, but its high-level reasoning ability remains lower than closed-source models. Llama 4 Scout, despite being lightweight, maintains stable performance on foundational tasks, though its accuracy decreases on high-level prediction tasks.

Across the board, open-source models tend to excel in low-level maneuver prediction (likely benefiting from large-scale video pretraining) but systematically fall short on complex multi-agent interaction and counterfactual prediction tasks. This performance gap is likely attributable to the scarcity of safety-critical and counterfactual scenarios in standard pretraining corpora, highlighting the value of \dataset as a targeted training resource for foresight-oriented supervision.

\textbf{Finetuning with \dataset.} We further finetune Qwen3-VL-8B using \dataset, yielding the model Qwen3-VL-8B-FI. 

Finetuning on \dataset significantly boosts Qwen3-VL-8B's performance 
within the nuScenes domain, allowing it to outperform all baselines---
demonstrating that \dataset provides targeted supervision for the 
foresight reasoning gap identified in zero-shot evaluations.
Cross-domain generalization (\eg, to Waymo or nuPlan\cite{caesar2021nuplan}) remains 
an important direction for future work.

\subsection{WM-oriented Evaluation}
We use VLM zero-shot performance as our baseline. 
The semantic coherence of WM-generated futures is then quantified 
by measuring downstream QA performance gains when VLMs are 
augmented with predicted video and/or trajectory.
Crucially, to validate that observed improvements genuinely stem from semantically meaningful WM content rather than incidental input augmentation, we additionally conduct a \textbf{shuffled control experiment}: for each test scene, the WM-predicted video and trajectory are replaced with predictions from a randomly selected \emph{different} scene. Shuffled predictions consistently \emph{decrease} VLM accuracy across all task categories relative to the baseline, confirming that the gains under ground-truth WM predictions arise from semantically informative future content, not from the mere presence of additional visual tokens.

The results using data generated by \textit{Epona} as input are shown in \cref{fig:performance on epona}, while the results using data generated by \textit{DrivingWorld} as input are shown in \cref{fig:performance on drivingworld}.

\textbf{Future Augmentation with Epona.}
When predicted video is incorporated (b), all models show noticeable gains in tasks requiring short-term motion anticipation. In particular, GPT-5 benefits most from video cues, especially in Relative Distance and Ego--VRU Relative Position, indicating that video sequences effectively support mid-level dynamic understanding. With only predicted trajectories as additional input (c), the improvements shift toward relational reasoning. Models such as Claude-Sonnet-4.5 show larger gains in Pedestrian Intent and Risk Area, suggesting trajectory inputs provide clearer signals for future agent--agent interactions. Meanwhile, Qwen models exhibit more conservative improvements, implying a stronger reliance on visual cues. Finally, combining both predicted video and trajectories (d) yields the most comprehensive benefits. The finetuned Qwen3-VL-8B-FI achieves the most significant and uniform boosts across nearly all tasks, surpassing larger models in several categories such as Counterfactual Prediction and Ego--VRU Relative Position. This indicates that multi-modal predictive inputs effectively complement its finetuned foresight capability. GPT-5 also reaches its strongest performance in this setting, particularly in dynamic maneuver understanding (Turn Change, Lane Change). Overall, these results demonstrate that (1) predictive modalities enhance foresight reasoning in complementary ways—video for motion cues, trajectories for relational cues; and (2) models trained on \dataset, even with smaller parameter sizes, can substantially benefit from predictive inputs and outperform much larger closed-source systems.

\textbf{Future Augmentation with DrivingWorld.}
DrivingWorld-generated predictions yield larger and more consistent 
gains across VLMs than Epona, particularly on tasks requiring 
temporal consistency and agent--agent relational reasoning (\eg, 
Relative Position, Pedestrian Intent). The video--trajectory fusion 
setting again achieves the strongest overall performance, with 
improvements in Risk Area and E-V Rel. Pos. being especially 
pronounced---suggesting that DrivingWorld's autoregressive architecture 
produces more geometrically coherent futures that better complement 
VLMs' semantic reasoning. In contrast, Epona's gains are more modest 
and task-selective, suggesting its predicted futures are less 
complementary to VLMs' existing visual reasoning, possibly due to 
differences in motion modeling fidelity or temporal resolution.

\section{Conclusion}
\label{sec:conclusion}

We presented \dataset and \bench for training and evaluating Foresight 
Intelligence in autonomous driving. Our evaluation across 13 VLMs reveals 
three findings: (1) current models systematically underperform on 
high-level counterfactual tasks, with the gap not explained by model 
scale alone; (2) world-model-generated futures provide architecture-dependent 
semantic gains, with DrivingWorld outperforming Epona on relational 
reasoning tasks; (3) fine-tuning on \dataset substantially closes the 
foresight gap within the nuScenes domain.
Future work should address cross-domain generalization and domain adaptation across driving datasets~\cite{pan2026panda}, as well as tighter integration between world model generation and VLM-based semantic reasoning.

\section*{Acknowledgments} 
This work is supported by the Agency for Science, Technology and Research (A*STAR) under its Career Development Fund (Project No. H26-KSR0066) and is also supported by the Agency for Science, Technology and Research (A*STAR) under its MTC Programmatic Funds (Grant No. M23L7b0021).

The authors would like to sincerely thank the Program Chairs, Area Chairs, and Reviewers for the time and efforts devoted during the review process.

% \clearpage

% ---- Bibliography ----
\bibliographystyle{splncs04}
\bibliography{main}

\clearpage

% ================================================================
%  PART 2: Supplementary Material (as Appendix)
% ================================================================
\appendix
\begin{center}
    \LARGE\bfseries Thinking Ahead: Foresight Intelligence in MLLMs and World Model\\[0.5em]
    \Large\bfseries Supplementary Material
\end{center}

\vspace{1em}

\section{Task Descriptions}
\label{sec:task_desc}

\subsection{Task Definition}

In this section, we provide detailed descriptions of each task defined in the dataset.

\subsubsection{Spatio-temporal Dynamic Reasoning (Low-level)}

\quad \textbf{Speed Change. } This task evaluates the model's capability to perceive and anticipate the longitudinal motion dynamics of the ego vehicle. Given historical visual observations and trajectories, the model is required to predict the trend of the ego vehicle's speed over a specified time window (\eg, the next 0\textasciitilde3s, 3\textasciitilde6s, or other intervals defined in the query). This is formulated as a classification problem, requiring the model to distinguish between acceleration, deceleration, or maintaining a constant speed. 

\textbf{Turn Change. } This task assesses the model’s ability to perceive and anticipate the lateral motion dynamics of the ego vehicle. While the Speed Change task focuses on longitudinal control, Turn Change requires the model to predict the steering behavior over a specified time window based on historical context. The prediction is classified into four distinct categories: turn left, turn right, keep straight, or U-turn. 

\textbf{Lane Change. } This task evaluates the model's ability to anticipate the ego vehicle's lane-changing behavior relative to the road topology. Distinct from the ``Turn Change'' task which relies on kinematic heading, ``Lane Change'' requires understanding the vehicle's position within the HD map. Given historical observations, the model must predict whether the ego vehicle will change lanes within a specified time window. 

\textbf{Relative Distance. } This task evaluates the model's ability to perceive spatial depth and dynamic interactions between agents. Specifically, it focuses on predicting the change in distance between the ego vehicle and the nearest moving vehicle (in the front view). Given the historical context, the model must determine how the relative distance evolves over a specified time window. 

\textbf{Relative Position. } This task focuses on the spatial directional relationship between the ego vehicle and the surrounding traffic. While ``Relative Distance'' estimates the depth, this task requires the model to categorize the direction of the nearest moving vehicle relative to the ego vehicle's heading at the end of a specified time window. The relative position is discretized into 8 directional regions: ahead, behind, right, left, right-front, left-front, left-rear and right-rear.

\subsubsection{VRU-centric Risk Assessment (Mid-level)}

\quad \textbf{Pedestrian Intent. } This task focuses on understanding the behavior of Vulnerable Road Users (VRUs). The model is required to identify the nearest pedestrian in the historical front view and predict their primary intention over the next 3 seconds. The intent is classified into five critical behavioral modes: waiting at curbside, jaywalking, crossing in crosswalk, walking parallel to road and other irrelevant. 

\textbf{Ego-VRU Relative Position. } This task is a specialized variant of the ``Relative Position'' task, explicitly focusing on Vulnerable Road Users such as pedestrians, cyclists, and motorcyclists. Given the historical context, the model must identify the nearest VRU and predict its spatial directional relationship relative to the ego vehicle at the end of 3 seconds. This task is critical for assessing safety risks, as VRUs often exhibit more flexible and unpredictable motion patterns than vehicles. 

\textbf{Risk Area. } This task evaluates the model's environmental awareness and its ability to anticipate potential hazards based on scene semantics. Instead of predicting immediate collisions, this task requires the model to classify the environmental risk level of the areas the ego vehicle will enter within a specified time window. The risk is categorized into three specific high-risk scenarios: VRU risk area, high occlusion area and complex intersection. 

\subsubsection{High-level Causal Reasoning (High-level)}

\quad \textbf{Counterfactual Prediction. } This task evaluates the model's causal reasoning capabilities and its sensitivity to safety-critical scenarios. Unlike standard prediction tasks that ask about the actual future, this task presents a hypothetical ``what-if'' query: ``If the ego vehicle were to perform action $A$ (\eg, turn left, accelerate) within a specified time window, what would be the most likely outcome?'' The model must assess whether this hypothetical maneuver would lead to a collision, a traffic violation, or a safe outcome.

\subsection{QA templates}

We present the prompt templates organized by their cognitive complexity. \cref{QA: Speed change,QA: Turn Change,QA: Lane Change,QA: Relative Distance,QA: Relative Position} detail the templates for Low-level Spatio-temporal tasks, \cref{QA: Pedestrian Intent,QA: Ego-VRU Relative Position,QA: Risk Area} for Mid-level Risk Assessment, and \cref{QA: Countefactual Prediction} for High-level Causal Reasoning.

\begin{figure*}[t]  
\centering
\begin{tcolorbox}[
    width=0.9\textwidth,    
    title={QA Template: Speed Change},
    colback=black!3,
    colframe=black!70,
    coltitle=white,
    fonttitle=\bfseries,
    arc=2mm,
    boxrule=0.6pt,
    left=3mm,
    right=3mm,
    top=0.5mm, bottom=0.5mm,
]
\raggedright\small\ttfamily

\textbf{Question:} You are an expert driving assistant. What is the speed change trend of the ego vehicle in the next \{\textit{interval}\} seconds? The answer must be one of:\\ A. Constant Speed\\
B. Acceleration\\
C. Deceleration\\
\textbf{Answer:} A/B/C
\end{tcolorbox}
\caption{\textbf{QA Template: Speed Change.}}
\label{QA: Speed change}
\end{figure*}

\begin{figure*}[t]  
\centering
\begin{tcolorbox}[
    width=0.9\textwidth,    
    title={QA Template: Turn Change},
    colback=black!3,
    colframe=black!70,
    coltitle=white,
    fonttitle=\bfseries,
    arc=2mm,
    boxrule=0.6pt,
    left=3mm,
    right=3mm,
    top=0.5mm, bottom=0.5mm,
]
\raggedright\small\ttfamily

\textbf{Question:} You are an expert driving assistant. What is the turn change trend of the ego vehicle in the next \{\textit{interval}\} seconds? The answer must be one of:\\ A. Straight\\
B. Left turn\\
C. Right turn\\
D. U-turn\\
\textbf{Answer:} A/B/C/D
\end{tcolorbox}
\caption{\textbf{QA Template: Turn Change}}
\label{QA: Turn Change}
\end{figure*}

\begin{figure*}[t]  
\centering
\begin{tcolorbox}[
    width=0.9\textwidth,    
    title={QA Template: Lane Change},
    colback=black!3,
    colframe=black!70,
    coltitle=white,
    fonttitle=\bfseries,
    arc=2mm,
    boxrule=0.6pt,
    left=3mm,
    right=3mm,
    top=0.5mm, bottom=0.5mm,
]
\raggedright\small\ttfamily

\textbf{Question:} You are an expert driving assistant. Will the ego vehicle change lanes in the next \{\textit{interval}\} seconds? The answer must be one of:\\
A. Left Lane Change\\
B. Right Lane Change\\
C. No Lane Change\\
\textbf{Answer:} A/B/C
\end{tcolorbox}
\caption{\textbf{QA Template: Lane Change.}}
\label{QA: Lane Change}
\end{figure*}

\begin{figure*}[t]  
\centering
\begin{tcolorbox}[
    width=0.9\textwidth,    
    title={QA Template: Relative Distance},
    colback=black!3,
    colframe=black!70,
    coltitle=white,
    fonttitle=\bfseries,
    arc=2mm,
    boxrule=0.6pt,
    left=3mm,
    right=3mm,
    top=0.5mm, bottom=0.5mm,
]
\raggedright\small\ttfamily

\textbf{Question:} You are an expert driving assistant. Focus on the nearest \{\textit{description}\} vehicle in front. How will the relative distance between your vehicle and this vehicle change over the next \{\textit{interval}\} seconds? The answer must be one of:\\
A. Approaching\\
B. Distancing\\
C. Maintaining\\
D. Approaching then distancing\\
\textbf{Answer:} A/B/C/D
\end{tcolorbox}
\caption{\textbf{QA Template: Relative Distance.}}
\label{QA: Relative Distance}
\end{figure*}

\begin{figure*}[ht!]  
\centering
\begin{tcolorbox}[
    width=0.9\textwidth,    
    title={QA Template: Relative Position},
    colback=black!3,
    colframe=black!70,
    coltitle=white,
    fonttitle=\bfseries,
    arc=2mm,
    boxrule=0.6pt,
    left=3mm,
    right=3mm,
    top=0.5mm, bottom=0.5mm,
]
\raggedright\small\ttfamily

\textbf{Question:} You are an expert driving assistant with localization capabilities. Focus on the nearest \{\textit{description}\} vehicle in front. Predict its relative position to the ego-vehicle over the next \{\textit{interval}\} seconds. The answer must be one of the following eight:\\
A. Ahead\\
B. Behind\\
C. Right\\
D. Left\\
E. Right-front\\
F. Left-front\\
G. Left-rear\\
H. Right-rear\\
\textbf{Answer:} A/B/C/D/E/F/G/H
\end{tcolorbox}
\caption{\textbf{QA Template: Relative Position.}}
\label{QA: Relative Position}
\end{figure*}

\begin{figure*}[t]  
\centering
\begin{tcolorbox}[
    width=0.9\textwidth,    
    title={QA Template: Pedestrian Intent},
    colback=black!3,
    colframe=black!70,
    coltitle=white,
    fonttitle=\bfseries,
    arc=2mm,
    boxrule=0.6pt,
    left=3mm,
    right=3mm,
    top=0.5mm, bottom=0.5mm,
]
\raggedright\small\ttfamily

\textbf{Question:} You are an expert driving assistant with a focus on pedestrian behavior. Observe the closest pedestrian in the front view. What is their primary intent over the next 3 seconds? The answer must be one of:\\
A. Waiting at Curbside\\
B. Jaywalking\\
C. Crossing in Crosswalk\\
D. Walking Parallel to Road\\
E. Other Irrelevant\\
\textbf{Answer:} A/B/C/D/E
\end{tcolorbox}
\caption{\textbf{QA Template: Pedestrian Intent.}}
\label{QA: Pedestrian Intent}
\end{figure*}

\begin{figure*}[t]  
\centering
\begin{tcolorbox}[
    width=0.9\textwidth,    
    title={QA Template: Ego-VRU Relative Position},
    colback=black!3,
    colframe=black!70,
    coltitle=white,
    fonttitle=\bfseries,
    arc=2mm,
    boxrule=0.6pt,
    left=3mm,
    right=3mm,
    top=0.5mm, bottom=0.5mm,
]
\raggedright\small\ttfamily

\textbf{Question:} You are an expert driving assistant with localization capabilities. Focus on the closest {VRU} in the front view. Predict its relative position to the ego-vehicle after 3 seconds. The answer must be one of the following eight:\\
A. Ahead\\
B. Behind\\
C. Right\\
D. Left\\
E. Right-front\\
F. Left-front\\
G. Left-rear\\
H. Right-rear\\
\textbf{Answer:} A/B/C/D/E/F/G/H
\end{tcolorbox}
\caption{\textbf{QA Template: Ego-VRU Relative Position.}}
\label{QA: Ego-VRU Relative Position}
\end{figure*}

\begin{figure*}[t]  
\centering
\begin{tcolorbox}[
    width=0.9\textwidth,    
    title={QA Template: Risk Area},
    colback=black!3,
    colframe=black!70,
    coltitle=white,
    fonttitle=\bfseries,
    arc=2mm,
    boxrule=0.6pt,
    left=3mm,
    right=3mm,
    top=0.5mm, bottom=0.5mm,
]
\raggedright\small\ttfamily

\textbf{Question:} You are an expert driving assistant with advanced situational awareness. Based on the current scene, which type(s) of high-risk area, if any, will the ego vehicle primarily enter within the next 3 seconds? Return the letter(s) corresponding to the selected option(s), separated by commas (\eg, `A, B'). The answer may be one or more of: \\
A. VRU Risk\\
B. High Occlusion Area\\
C. Complex Intersection\\
D. No Risk Area\\ 
\textbf{Answer:} A, B (multiple correct answers)
\end{tcolorbox}
\caption{\textbf{QA Template: Risk Area.}}
\label{QA: Risk Area}
\end{figure*}

\begin{figure*}[t]  
\centering
\begin{tcolorbox}[
    width=0.9\textwidth,    
    title={QA Template: Counterfactual Prediction},
    colback=black!3,
    colframe=black!70,
    coltitle=white,
    fonttitle=\bfseries,
    arc=2mm,
    boxrule=0.6pt,
    left=3mm,
    right=3mm,
    top=0.5mm, bottom=0.5mm,
]
\raggedright\small\ttfamily

\textbf{Question:} You are an expert driving assistant with predictive capabilities. If the ego vehicle were to \{\textit{act}\} over the next 3 seconds, what would be the most likely outcome? This is a multiple-choice question, and the answer can be one or more of:\\
A. Collision\\
B. Driving Out of Legal Area\\
C. Violating Traffic Rules\\
D. Safe\\
\textbf{Answer:} A, B, C (multiple correct answers)
\end{tcolorbox}
\caption{\textbf{QA Template: Counterfactual Prediction.}}
\label{QA: Countefactual Prediction}
\end{figure*}

\subsection{Evaluation Prompt}
We employed different prompts to evaluate the performance of VLMs and WMs under varied input conditions (see \cref{prompt: baseline,prompt: base + pred. video,prompt: base + pred. traj,prompt: base & pred. v+t}).

\begin{figure*}[t]  
\centering
\begin{tcolorbox}[
    width=0.9\textwidth,    
    title={Prompt: Baseline},
    colback=black!3,
    colframe=black!70,
    coltitle=white,
    fonttitle=\bfseries,
    arc=2mm,
    boxrule=0.6pt,
    left=3mm,
    right=3mm,
    top=0.5mm, bottom=0.5mm,
]
\raggedright\small\ttfamily
Trajectory 1: [x, y, yaw]\\
$\vdots$\\
Trajectory 7: [x, y, yaw]\\
Frame 1: \{IMAGE\_TOKEN\}\\
$\vdots$\\
Frame 7: \{IMAGE\_TOKEN\}\\
The input image is captured by the vehicle's front-facing camera within 3 seconds.\\
The pose of each trajectory point corresponds to each image frame in sequence.\\
The trajectory points are relative coordinates, with the trajectory point of the last frame as the origin; the yaw angles are also relative values, with the yaw angle of the last frame as the reference.\\
Based on the input, Please answer the following questions in order. You only need to answer with the option number, without providing the specific content of the option.\\
Number each answer according to the question:\\
\ldots\\
Definitions:\\
VRU\_Risk means there is a vulnerable road user in the field of view that warrants attention.\\
High\_Occlusion\_Area means the field of view will be highly occluded by large vehicles.\\
Complex\_Intersection means the ego vehicle will pass through an intersection that contains at least three moving vehicles.
\end{tcolorbox}
\caption{\textbf{Prompt: Baseline.}}
\label{prompt: baseline}
\end{figure*}

\begin{figure*}[t]  
\centering
\begin{tcolorbox}[
    width=0.9\textwidth,    
    title={Prompt: Base \& Pred. Video},
    colback=black!3,
    colframe=black!70,
    coltitle=white,
    fonttitle=\bfseries,
    arc=2mm,
    boxrule=0.6pt,
    left=3mm,
    right=3mm,
    top=0.5mm, bottom=0.5mm,
]
\raggedright\small\ttfamily
Trajectory 1: [x, y, yaw]\\
$\vdots$\\
Trajectory 7: [x, y, yaw]\\
Historical Frame 1: \{IMAGE\_TOKEN\}\\
$\vdots$\\
Historical Frame 7: \{IMAGE\_TOKEN\}\\
Predicted Frame 1: \{IMAGE\_TOKEN\}\\
$\vdots$\\
Predicted Frame 6: \{IMAGE\_TOKEN\}\\
You are given two types of images:\\
1. Historical frames: 7 images captured by the vehicle's front-facing camera over 3 seconds. Each image corresponds to a historical trajectory point, in chronological order.\\
2. Predicted future frames: 6 images generated by a world model, representing possible future views of the vehicle in the next 3 seconds after the last historical frame.\\
The current moment is defined as the last historical frame. All trajectory points are historical and correspond to each historical frame.\\
Trajectory points are provided as relative coordinates and yaw angles (x, y, radians), all relative to the last historical frame.\\
The predicted future frames are provided to assist you in answering the following questions about possible future situations.\\
Based on the input, please answer the following questions in order. You MUST ONLY reply with the option number for each question. Number your answers according to the question order: \\
\ldots\\
Definitions:\\
VRU\_Risk: A vulnerable road user is present and requires attention.\\
High\_Occlusion\_Area: The view will be highly occluded by large vehicles.\\
Complex\_Intersection: The ego vehicle will pass through an intersection with at least three moving vehicles.
\end{tcolorbox}
\caption{\textbf{Prompt: Base \& Pred. Video.}}
\label{prompt: base + pred. video}
\end{figure*}

\begin{figure*}[t]  
\centering
\begin{tcolorbox}[
    width=0.9\textwidth,    
    title={Prompt: Base \& Pred. Traj},
    colback=black!3,
    colframe=black!70,
    coltitle=white,
    fonttitle=\bfseries,
    arc=2mm,
    boxrule=0.6pt,
    left=3mm,
    right=3mm,
    top=0.5mm, bottom=0.5mm,
]
\raggedright\small\ttfamily
Historical Trajectory 1: [x, y, yaw]\\
$\vdots$\\
Historical Trajectory 7: [x, y, yaw]\\
Historical Frame 1: \{IMAGE\_TOKEN\}\\
$\vdots$\\
Historical Frame 7: \{IMAGE\_TOKEN\}\\
Predicted Trajectory 1: [x, y, yaw]\\
$\vdots$\\
Predicted Trajectory \textit{x}: [x, y, yaw] (\textit{x} depending on outputs of WMs)\\
You are given three types of input:\\
1. Historical frames: 7 images captured by the vehicle's front-facing camera over 3 seconds. Each image corresponds to a historical trajectory point, in chronological order.\\
2. Historical trajectory: 7 trajectory points (relative coordinates and yaw angles) corresponding to the historical frames, all relative to the last historical frame.\\
3. Predicted trajectory: \textit{x} trajectory points (relative coordinates and yaw angles) predicted for the next 3 seconds after the last historical frame.\\
The current moment is defined as the last historical frame. All trajectory points are relative to the last historical frame.\\
The predicted trajectory is provided to assist you in answering the following questions about possible future situations within the next 3 seconds.\\
Based on the input, you must answer the following questions in order. You can't refuse to answer! You MUST ONLY reply with the option number for each question. Number your answers according to the question order:\\
\ldots\\
Definitions:\\
VRU\_Risk: A vulnerable road user is present and requires attention.\\
High\_Occlusion\_Area: The view will be highly occluded by large vehicles.\\
Complex\_Intersection: The ego vehicle will pass through an intersection with at least three moving vehicles.
\end{tcolorbox}
\caption{\textbf{Prompt: Base \& Pred. Traj.}}
\label{prompt: base + pred. traj}
\end{figure*}

\begin{figure*}[t]  
\centering
\begin{tcolorbox}[
    width=0.9\textwidth,    
    title={Prompt: Base \& Pred. V+T},
    colback=black!3,
    colframe=black!70,
    coltitle=white,
    fonttitle=\bfseries,
    arc=2mm,
    boxrule=0.6pt,
    left=3mm,
    right=3mm,
    top=0.5mm, bottom=0.5mm,
]
\raggedright\small\ttfamily
Historical Trajectory 1: [x, y, yaw]\\
$\vdots$\\
Historical Trajectory 7: [x, y, yaw]\\
Historical Frame 1: \{IMAGE\_TOKEN\}\\
$\vdots$\\
Historical Frame 7: \{IMAGE\_TOKEN\}\\
Predicted Trajectory 1: [x, y, yaw]\\
$\vdots$\\
Predicted Trajectory \textit{x}: [x, y, yaw] (\textit{x} depending on outputs of WMs)\\
Predicted Frame 1: \{IMAGE\_TOKEN\}\\
$\vdots$\\
Predicted Frame 6: \{IMAGE\_TOKEN\}\\
You are given four types of input:\\
1. Historical frames: 7 images captured by the vehicle's front-facing camera over 3 seconds. Each image corresponds to a historical trajectory point, in chronological order.\\
2. Historical trajectory: 7 trajectory points (relative coordinates and yaw angles) corresponding to the historical frames, all relative to the last historical frame.\\
3. Predicted future frames: 6 images generated by a world model, representing possible future views of the vehicle in the next 3 seconds after the last historical frame.\\
4. Predicted trajectory: \textit{x} trajectory points (relative coordinates and yaw angles) predicted for the next 3 seconds after the last historical frame. These points do not necessarily correspond to the predicted frames one by one.\\
The current moment is defined as the last historical frame. All historical trajectory points are relative to the last historical frame.\\
The predicted future frames and predicted trajectory are provided to assist you in answering the following questions about possible future situations within the next 3 seconds.\\
Based on the input, you must answer the following questions in order.You can't refuse to answer! You MUST ONLY reply with the option number for each question. Number your answers according to the question order:\\
\ldots\\
Definitions:\\
VRU\_Risk: A vulnerable road user is present and requires attention.\\
High\_Occlusion\_Area: The view will be highly occluded by large vehicles.\\
Complex\_Intersection: The ego vehicle will pass through an intersection with at least three moving vehicles.
\end{tcolorbox}
\caption{\textbf{Prompt: Base \& Pred. V+T.}}
\label{prompt: base & pred. v+t}
\end{figure*}

\section{Implementation Details}

\subsection{Data Filtering}
To ensure that every question in \dataset is valid and answerable, we implement a strict Task-Specific Filtering mechanism. We do not apply a generic filter; instead, we evaluate each frame against specific entry conditions for each of the nine tasks. A question is generated for a specific task only if the scene satisfies the necessary prerequisites for that task.

\subsection{Experiment Details}
\textbf{Hyperparameter Settings for Inference.} To ensure a fair and robust evaluation, we adopted specific inference strategies tailored to the nature of the models.

API-based Models: This category includes both proprietary closed-source models (\eg, GPT-5, Claude-Sonnet-4.5, Gemini-2.5) and hosted open-source models accessed via APIs (\eg, Llama-4-Scout). For these models, we strictly adhered to the default settings provided by their respective official APIs. This approach ensures that we evaluate the representative ``out-of-the-box'' capabilities of these models as they are typically used by end-users and developers.

Locally Deployed Models: This category consists of the locally deployed Qwen series, with a precision of BF16. For these models, we utilized a low-temperature sampling strategy to balance determinism with generation stability. Specifically, we set the temperature to 0.1. This setting minimizes randomness to ensure consistent answers for scientific evaluation while retaining a slight probabilistic margin to prevent repetition loops often observed in strict greedy decoding modes.

\textbf{Hyperparameter Settings for Finetuning.} For the finetuned model Qwen3-VL-8B-FI, we performed full parameter fine-tuning on the LLM backbone and the vision-language projector, while keeping the vision encoder frozen. To manage the memory footprint of the 8B model with long context windows, we utilized DeepSpeed ZeRO-3 optimization. The detailed training configuration is listed in \cref{tab:finetune_hyperparams}.

\subsection{Frequency Alignment}
To effectively bridge the gap between high-frequency sensor data and the token limitations of VLMs, we implement a unified frequency alignment strategy.

\textbf{Historical Frames.} The raw visual data from the nuScenes dataset is recorded at 12 Hz. Directly feeding this dense stream into VLMs is computationally prohibitive. Therefore, we downsample the 3-second historical window to 2 Hz. Crucially, to fully capture the temporal boundaries of the observation window, we include both the start and end timestamps. This results in a sequence of 7 frames serving as the visual input for VLMs. This sampling strategy ensures that the model receives complete boundary information.

\textbf{Future frames.} Consistent with the input, the future frames are also sampled at 2 Hz for evaluation. We note that the native generation frequency of future frames depends on the World Model being evaluated. To ensure a standardized evaluation on \bench, we synchronize their generated outputs by sampling them to the same 2 Hz frequency before feeding them into the VLM.

\section{Additional Quantitative Results}
The joint evaluation results of \textit{Epona} and VLMs are presented in \cref{tab:Base + Pred. Video(Epona) eval}, \cref{tab:Base + Pred. Traj(Epona) eval} and \cref{tab:Base + Pred. V+T(Epona) eval}. While the joint evaluation results of \textit{DrivingWorld} and VLMs are presented in \cref{tab:Base + Pred. Video(DW) eval}, \cref{tab:Base + Pred. Traj(DW) eval} and \cref{tab:Base + Pred. V+T(DW) eval}.

\begin{table*}[ht!]
    \centering
    \resizebox{0.9\linewidth}{!}{ 
        \begin{tabular}{l|l|l}
            \toprule
            \textbf{Hyperparameter} & \textbf{Value} & \textbf{Description} \\
            \midrule
            Base Model & Qwen3-VL-8B-Instruct & Initial checkpoint initialized from HuggingFace \\
            Tuned Modules & LLM + Projector & Vision encoder is frozen (\texttt{tune\_mm\_vision=False}) \\
            Optimization Strategy & DeepSpeed ZeRO-3 & Memory optimization for full fine-tuning \\
            Precision & BF16 & Bfloat16 mixed precision training \\
            \midrule
            Learning Rate & $2e-5$ & Peak learning rate \\
            LR Scheduler & Cosine & Cosine decay with warmup \\
            Warmup Ratio & 0.05 & Ratio of steps used for learning rate warmup \\
            Weight Decay & 0.01 & Regularization parameter \\
            \midrule
            Num Epochs & 2 & Total number of training epochs \\
            Batch Size & 32 (Global) & 1 per device $\times$ 4 accum steps $\times$ 8 GPUs \\
            Max Sequence Length & 6000 & To accommodate long video context \\
            Computational Resources & 8 $\times$ NVIDIA A40 & 48GB VRAM per GPU \\
            \bottomrule
        \end{tabular}
    }
    \caption{\textbf{Hyperparameters for Finetuning Qwen3-VL-8B on \dataset.} We perform full parameter fine-tuning on the LLM backbone and projector while keeping the vision tower frozen.}
    \label{tab:finetune_hyperparams}
\end{table*}

\begin{figure*}[t!]
    \captionsetup{type=table}
    \vspace{-0.4cm}
    \centering
    \resizebox{0.9\textwidth}{!}{
    \begin{tabular}{r|cc|ccccccccc}
    & & &
    \rotatebox{75}{Speed Change} &
    \rotatebox{75}{Turn Change} &
    \rotatebox{75}{Lane Change} &
    \rotatebox{75}{Rel. Dist.} & 
    \rotatebox{75}{Rel. Pos.} &
    \rotatebox{75}{Ped. Int.} &
    \rotatebox{75}{E-V Rel. Pos.} &
    \rotatebox{75}{Risk Area} &
    \rotatebox{75}{CFP} \\
    Methods & Rank & Overall & \multicolumn{5}{c}{\cellcolor{red!10}Low-level} & \multicolumn{3}{c}{\cellcolor{orange!10}Mid-level} & \multicolumn{1}{c}
    {\cellcolor{yellow!10}High-level}\\
    \hline
    \rowcolor{navyblue!5}
    \multicolumn{1}{l|}{\textcolor{black}{\textit{Open-source Models}}} & & & & & & & & & & & \\
    Qwen2.5-VL-72B &5&47.30&27.33&\cellcolor{oai-gray-600}{92.33}&97.50&36.18&15.65&27.54&7.41&64.00&17.80 \\
    Qwen2.5-VL-7B  &10&43.38&25.33&92.33&97.50&27.64&14.43&42.03&3.70&46.00&8.43 \\
    Qwen3-VL-32B &\cellcolor{oai-green-200}{3}&47.45&30.00&91.33&96.67&35.16&14.84&37.68&29.63&63.33&15.93 \\
    Qwen3-VL-8B &11&43.17&29.67&87.17&92.17&35.37&14.63&39.13&9.88&60.67&3.75\\
    \hline
    \rowcolor{navyblue!5}
    \multicolumn{1}{l|}{\textcolor{black}{\textit{Closed-source Models}}} & & & & & & & & & & & \\
    GPT-4o Mini &9&43.75&32.00&88.33&97.50&33.94&14.43&18.84&4.94&7.33&13.79 \\
    GPT-5 &\cellcolor{oai-green-400}{2}&47.87&32.67&77.33&89.17&\cellcolor{oai-gray-600}{36.99}&31.91&\cellcolor{oai-gray-600}{43.48}&51.85&49.33&\cellcolor{oai-gray-600}{20.62}\\
    Gemini-2.0 Flash &8&46.12&33.67&91.33&97.50&36.99&14.43&26.09&16.05&50.00&9.77 \\
    Gemini-2.5 Flash &7&46.57&\cellcolor{oai-gray-600}{40.00}&70.67&96.67&32.11&21.95&27.54&28.40&52.00&20.62 \\
    Gemini-2.5 Pro &6&47.14&37.17&83.50&97.00&28.86&\cellcolor{oai-gray-600}{21.95}&36.23&\cellcolor{oai-gray-600}{38.27}&34.00&19.14 \\
    Claude-3.7-Sonnet &\cellcolor{oai-green-200}{3}&47.45&33.33&91.17&97.33&35.16&17.89&28.99&17.28&45.33&16.60 \\
    Claude-Sonnet-4.5 &\cellcolor{oai-green-600}{1}&48.39&29.00&87.67&\cellcolor{oai-gray-600}{97.50}&34.55&21.54&33.33&24.69&\cellcolor{oai-gray-600}{68.00}&19.81 \\
    \hline
    \rowcolor{navyblue!5}
    \multicolumn{1}{l|}{\textcolor{black}{\textit{Finetuned Model}}}& & & & & & & & & & & \\
    Qwen3-VL-8B-FI &-&65.81&44.67&92.00&97.50&46.54&48.98&42.03&53.09&71.33&62.52 \\
    \end{tabular}
 }
    \caption{\textbf{Baseline \& Pred. Video (from Epona) Evaluation on \bench:} Input historical frames, trajectories and predicted frames generated from Epona. \colorbox{oai-gray-600}{Dark gray} cells indicate the best result across all models. }
    \label{tab:Base + Pred. Video(Epona) eval}
    \vspace{-0.4cm}
\end{figure*}

\begin{figure*}[t!]
    \captionsetup{type=table}
    \vspace{-0.4cm}
    \centering
    \resizebox{0.9\textwidth}{!}{
    \begin{tabular}{r|cc|ccccccccc}
    & & &
    \rotatebox{75}{Speed Change} &
    \rotatebox{75}{Turn Change} &
    \rotatebox{75}{Lane Change} &
    \rotatebox{75}{Rel. Dist.} & 
    \rotatebox{75}{Rel. Pos.} &
    \rotatebox{75}{Ped. Int.} &
    \rotatebox{75}{E-V Rel. Pos.} &
    \rotatebox{75}{Risk Area} &
    \rotatebox{75}{CFP} \\
    Methods & Rank & Overall & \multicolumn{5}{c}{\cellcolor{red!10}Low-level} & \multicolumn{3}{c}{\cellcolor{orange!10}Mid-level} & \multicolumn{1}{c}
    {\cellcolor{yellow!10}High-level}\\
    \hline
    \rowcolor{navyblue!5}
    \multicolumn{1}{l|}{\textcolor{black}{\textit{Open-source Models}}} & & & & & & & & & & & \\
    Qwen2.5-VL-72B &\cellcolor{oai-green-600}{1}&45.34 &35.00 &85.00 &96.67 &30.89 &14.43 &24.64 &8.64 &58.00 &13.79 \\
    Qwen2.5-VL-7B  & 8&42.10 &29.33 &83.00 &97.00 &27.64 &14.43 &34.78 &3.70 &40.00 &8.43 \\
    Qwen3-VL-32B & 9&39.31&41.67 &42.33 &95.83 &31.10 &14.63 & 35.78&28.40 &60.67 &8.57 \\
    Qwen3-VL-8B & 7&43.02 &33.50 &82.83 &94.00 &32.72 &14.43 &34.78 &7.41 &58.00 &4.95\\
    \hline
    \rowcolor{navyblue!5}
    \multicolumn{1}{l|}{\textcolor{black}{\textit{Closed-source Models}}} & & & & & & & & & & & \\
    GPT-4o Mini & 4& 43.57& 34.50& \cellcolor{oai-gray-600}{86.83}& 96.50& 34.35& 14.43& 20.29& 3.70& 4.00&13.25\\
    GPT-5 & 6&43.11 &26.66 &57.45 &93.05 & \cellcolor{oai-gray-600}{34.88}& \cellcolor{oai-gray-600}{25.81}& 36.23& 45.12& 45.70&\cellcolor{oai-gray-600}{21.41}\\
    Gemini-2.0 Flash & 11&36.41 &41.17 &25.67 &96.33 &33.33&15.45 &23.19 &14.81 &38.00 &12.18 \\
    Gemini-2.5 Flash & 10&38.76 &\cellcolor{oai-gray-600}{37.76} &31.50 &96.67 &32.52 &16.06 &26.09 &30.86 &48.00 &18.21\\
    Gemini-2.5 Pro & 5&43.33 &36.33 &65.00 &96.33 &29.27 &19.11 &\cellcolor{oai-gray-600}{37.68} &\cellcolor{oai-gray-600}{38.51} &30.00 &17.80\\
    Claude-3.7-Sonnet &\cellcolor{oai-green-200}{3}&43.62 &32.17 &72.50 &97.17&34.35 &15.45 &30.43 &17.28 &35.33 &17.00\\
    Claude-Sonnet-4.5 &\cellcolor{oai-green-400}{2}&44.40&37.00 &52.33&\cellcolor{oai-gray-600}{97.17}&33.94&25.00&34.78 &22.22& \cellcolor{oai-gray-600}{64.67}& 20.48\\
    \hline
    \rowcolor{navyblue!5}
    \multicolumn{1}{l|}{\textcolor{black}{\textit{Finetuned Model}}}& & & & & & & & & & & \\
    Qwen3-VL-8B-FI &-&66.38 &44.17 &92.33&97.17 &42.68 &50.00 &50.72 &61.73 &78.00 &64.66\\
    \end{tabular}
}
    \caption{\textbf{Baseline \& Pred. Traj (from Epona) Evaluation on \bench:} Input historical frames, trajectories and predicted trajectories generated from Epona. \colorbox{oai-gray-600}{Dark gray} cells indicate the best result across all models. }
    \label{tab:Base + Pred. Traj(Epona) eval}
    \vspace{-0.4cm}
\end{figure*}

\begin{figure*}[t!]
    \captionsetup{type=table}
    \vspace{-0.4cm}
    \centering
    \resizebox{0.9\textwidth}{!}{
    \begin{tabular}{r|cc|ccccccccc}
    & & &
    \rotatebox{75}{Speed Change} &
    \rotatebox{75}{Turn Change} &
    \rotatebox{75}{Lane Change} &
    \rotatebox{75}{Rel. Dist.} & 
    \rotatebox{75}{Rel. Pos.} &
    \rotatebox{75}{Ped. Int.} &
    \rotatebox{75}{E-V Rel. Pos.} &
    \rotatebox{75}{Risk Area} &
    \rotatebox{75}{CFP} \\
    Methods & Rank & Overall & \multicolumn{5}{c}{\cellcolor{red!10}Low-level} & \multicolumn{3}{c}{\cellcolor{orange!10}Mid-level} & \multicolumn{1}{c}
    {\cellcolor{yellow!10}High-level}\\
    \hline
    \rowcolor{navyblue!5}
    \multicolumn{1}{l|}{\textcolor{black}{\textit{Open-source Models}}} & & & & & & & & & & & \\
    Qwen2.5-VL-72B &\cellcolor{oai-green-600}{1} &46.83 &32.00 &90.50 &96.67 &32.93 &15.24 &26.09 &11.11 &\cellcolor{oai-gray-600}{64.00} &15.93 \\
    Qwen2.5-VL-7B  &6&43.23 &26.33 &\cellcolor{oai-gray-600}{92.17} &97.17 &27.64 &14.43 &36.23 &3.70 &42.67 &8.43 \\
    Qwen3-VL-32B & 9&40.98 &35.17 &60.50 &91.83 &30.49 &15.24 &33.33 &33.33 &60.67 &10.58 \\
    Qwen3-VL-8B &5&43.77 &33.50&85.17 &92.67 &33.54 &14.43 &36.23 &9.88 &60.00 &6.69\\
    \hline
    \rowcolor{navyblue!5}
    \multicolumn{1}{l|}{\textcolor{black}{\textit{Closed-source Models}}} & & & & & & & & & & & \\
    GPT-4o Mini & 8&42.86&30.00&84.83 &96.50 &36.18 &14.43 &18.84 &3.70 &4.67 &13.65\\
    GPT-5 &\cellcolor{oai-green-200}{3}&45.42 &29.50 &60.17 &96.17 &35.37 &\cellcolor{oai-gray-600}{30.89} &\cellcolor{oai-gray-600}{42.03} &\cellcolor{oai-gray-600}{\cellcolor{oai-gray-600}{48.15}} &50.67 &\cellcolor{oai-gray-600}{20.75}\\
    Gemini-2.0 Flash & 11&39.31 &\cellcolor{oai-gray-600}{42.67} &44.00 &96.67 &32.52 &16.87 &26.09 &11.11 &39.33 &10.31 \\
    Gemini-2.5 Flash &10 &40.30 &41.33 &35.00 &96.17 &34.55 &14.63 &23.19 &23.46 &60.00 &19.01\\
    Gemini-2.5 Pro &7&43.20 &33.50 &64.17 &96.00 &35.98 &18.70 &30.43 &38.27 &28.67 &17.27\\
    Claude-3.7-Sonnet &\cellcolor{oai-green-400}{2}&45.71 &31.50 &84.50 &97.00 &35.77 &18.23 &30.43 &16.05 &42.67 &14.59\\
    Claude-Sonnet-4.5 &4&45.03 &32.33 &67.33 &\cellcolor{oai-gray-600}{92.17} &\cellcolor{oai-gray-600}{36.79} &24.54 &28.99 &20.99 &56.67 &18.07\\
    \hline
    \rowcolor{navyblue!5}
    \multicolumn{1}{l|}{\textcolor{black}{\textit{Finetuned Model}}}& & & & & & & & & & & \\
    Qwen3-VL-8B-FI &-&65.23 &43.83 &92.17 &97.17 &44.83 &48.58 &37.68 &50.62 &71.33 &62.38\\
    \end{tabular}
 }
    \caption{\textbf{Baseline \& Pred. V+T (from Epona) Evaluation on \bench:} Input historical frames, trajectories, predicted frames and trajectories generated from Epona. \colorbox{oai-gray-600}{Dark gray} cells indicate the best result across all models. }
    \label{tab:Base + Pred. V+T(Epona) eval}
    \vspace{-0.4cm}
\end{figure*}

\begin{figure*}[t!]
    \captionsetup{type=table}
    \vspace{-0.4cm}
    \centering
    \resizebox{0.9\textwidth}{!}{
    \begin{tabular}{r|cc|ccccccccc}
    & & &
    \rotatebox{75}{Speed Change} &
    \rotatebox{75}{Turn Change} &
    \rotatebox{75}{Lane Change} &
    \rotatebox{75}{Rel. Dist.} & 
    \rotatebox{75}{Rel. Pos.} &
    \rotatebox{75}{Ped. Int.} &
    \rotatebox{75}{E-V Rel. Pos.} &
    \rotatebox{75}{Risk Area} &
    \rotatebox{75}{CFP} \\
    Methods & Rank & Overall & \multicolumn{5}{c}{\cellcolor{red!10}Low-level} & \multicolumn{3}{c}{\cellcolor{orange!10}Mid-level} & \multicolumn{1}{c}
    {\cellcolor{yellow!10}High-level}\\
    \hline
    \rowcolor{navyblue!5}
    \multicolumn{1}{l|}{\textcolor{black}{\textit{Open-source Models}}} & & & & & & & & & & & \\
    Qwen2.5-VL-72B &\cellcolor{oai-green-600}{1} &47.17 &33.00 &91.67 &96.33 &33.13 &15.04 &24.64 &6.17 &\cellcolor{oai-gray-600}{57.33} &18.21 \\
    Qwen2.5-VL-7B  &10 &43.12 &25.17 &\cellcolor{oai-gray-600}{92.17} &97.17 &27.64 &14.43 &37.68 &3.70 &44.00 &8.43 \\
    Qwen3-VL-32B & \cellcolor{oai-green-200}{3}&45.63 &29.00 &88.17 &95.00 &35.57 &14.84 &36.23 &23.46 &52.67 &13.92 \\
    Qwen3-VL-8B & 11&41.06 &30.17 &80.33 &88.50 &31.50 &14.63 &36.23 &11.11 &57.33 &4.28\\
    \hline
    \rowcolor{navyblue!5}
    \multicolumn{1}{l|}{\textcolor{black}{\textit{Closed-source Models}}} & & & & & & & & & & & \\
    GPT-4o Mini & 9&43.30 &32.50 &90.17 &97.17 &28.66 &14.43 &18.84 &3.70 &7.33 &13.52\\
    GPT-5 & 5&45.39 &32.33 &65.00 &95.50 &36.79 &\cellcolor{oai-gray-600}{26.22} &\cellcolor{oai-gray-600}{39.13} &\cellcolor{oai-gray-600}{50.62} &44.00 &18.47\\
    Gemini-2.0 Flash &7 &44.22 &29.50 &87.17 &94.67 &\cellcolor{oai-gray-600}{38.21} &16.46 &18.84 &4.94 &44.67 &9.77 \\
    Gemini-2.5 Flash &8 &43.64 &\cellcolor{oai-gray-600}{35.00} &63.83 &95.83 &30.69 &18.29 &36.23 &23.46 &49.33 &19.41\\
    Gemini-2.5 Pro & 6&44.69 &34.67 &71.50 &95.67 &32.52 &20.73 &31.88 &30.86 &26.67 &20.35\\
    Claude-3.7-Sonnet & 4&45.58 &30.00 &90.00 &97.17 &32.93 &15.45 &24.64 &18.52 &33.33 &16.47\\
    Claude-Sonnet-4.5 &\cellcolor{oai-green-400}{2} &47.01 &30.50 &81.33 &\cellcolor{oai-gray-600}{97.17} &36.59 &20.53 &24.64 &17.28 &55.33 &\cellcolor{oai-gray-600}{20.35}\\
    \hline
    \rowcolor{navyblue!5}
    \multicolumn{1}{l|}{\textcolor{black}{\textit{Finetuned Model}}}& & & & & & & & & & & \\
    Qwen3-VL-8B-FI & -&60.35 &40.00 &91.83 &97.17 &42.28 &30.08 &43.48 &48.15 &58.67 &56.89\\
    \end{tabular}
 }
    \caption{\textbf{Baseline \& Pred. Video (from DrivingWorld) Evaluation on \bench:} Input historical frames, trajectories and predicted frames generated from DrivingWorld. \colorbox{oai-gray-600}{Dark gray} cells indicate the best result across all models. }
    \label{tab:Base + Pred. Video(DW) eval}
    \vspace{-0.4cm}
\end{figure*}

\begin{figure*}[t!]
    \captionsetup{type=table}
    \vspace{-0.4cm}
    \centering
    \resizebox{0.9\textwidth}{!}{
    \begin{tabular}{r|cc|ccccccccc}
    & & &
    \rotatebox{75}{Speed Change} &
    \rotatebox{75}{Turn Change} &
    \rotatebox{75}{Lane Change} &
    \rotatebox{75}{Rel. Dist.} & 
    \rotatebox{75}{Rel. Pos.} &
    \rotatebox{75}{Ped. Int.} &
    \rotatebox{75}{E-V Rel. Pos.} &
    \rotatebox{75}{Risk Area} &
    \rotatebox{75}{CFP} \\
    Methods & Rank & Overall & \multicolumn{5}{c}{\cellcolor{red!10}Low-level} & \multicolumn{3}{c}{\cellcolor{orange!10}Mid-level} & \multicolumn{1}{c}
    {\cellcolor{yellow!10}High-level}\\
    \hline
    \rowcolor{navyblue!5}
    \multicolumn{1}{l|}{\textcolor{black}{\textit{Open-source Models}}} & & & & & & & & & & & \\
    Qwen2.5-VL-72B & 8&43.04 &37.83 &67.33 &88.50 &33.13 &18.29 &28.99 &14.81 &54.67 &16.06 \\
    Qwen2.5-VL-7B  & 7&43.25 &26.67 &\cellcolor{oai-gray-600}{90.33} &97.17 &27.64 &14.43 &37.68 &3.70 &48.67 &8.43 \\
    Qwen3-VL-32B & 6&43.49 &\cellcolor{oai-gray-600}{38.50} &73.33 &96.83 &29.67 &14.63 &31.88 &28.40 &61.33 &7.90 \\
    Qwen3-VL-8B & 9&42.00 &29.83 &77.83 &91.33 &33.33 &14.63 &36.23 &8.64 &56.67 &8.30\\
    \hline
    \rowcolor{navyblue!5}
    \multicolumn{1}{l|}{\textcolor{black}{\textit{Closed-source Models}}} & & & & & & & & & & & \\
    GPT-4o Mini & 5&43.77&34.50&88.83 &97.00 &31.50 &14.43 &18.84 &4.94 &5.33 &13.92\\
    GPT-5 & \cellcolor{oai-green-600}{1}&48.11 &30.67 &71.00 &96.67 &\cellcolor{oai-gray-600}{39.63} &\cellcolor{oai-gray-600}{30.89} &\cellcolor{oai-gray-600}{39.13} &\cellcolor{oai-gray-600}{49.38} &52.67 &21.42\\
    Gemini-2.0 Flash &11 &38.32 &32.17 &48.00 &95.33 &34.15 &16.26 &26.09 &16.05 &37.33 &10.71 \\
    Gemini-2.5 Flash & 10&40.80 &32.00 &46.33 &97.17 &32.32 &20.73 &31.88 &30.86 &50.67 &16.87\\
    Gemini-2.5 Pro & 4&44.34 &30.57 &69.26 &93.92 &33.68 &23.14 &38.81 &29.11 &39.19 &19.67\\
    Claude-3.7-Sonnet &\cellcolor{oai-green-200}{3} &46.10 &28.83 &86.83 &\cellcolor{oai-gray-600}{97.17} &33.74 &16.87 &34.78 &22.22 &38.67 &18.74\\
    Claude-Sonnet-4.5 &\cellcolor{oai-green-400}{2}&46.96 &32.50 &70.67 &96.17 &37.80 &23.17 &30.43 &18.52 &\cellcolor{oai-gray-600}{64.67} &\cellcolor{oai-gray-600}{22.76}\\
    \hline
    \rowcolor{navyblue!5}
    \multicolumn{1}{l|}{\textcolor{black}{\textit{Finetuned Model}}}& & & & & & & & & & & \\
    Qwen3-VL-8B-FI &-&67.11 &46.00 &92.33 &97.17 &45.33 &51.02 &50.72 &59.26 &78.67 &64.66\\
    \end{tabular}
 }
    \caption{\textbf{Baseline \& Pred. Traj (from DrivingWorld) Evaluation on \bench:} Input historical frames, trajectories and predicted trajectories generated from DrivingWorld. \colorbox{oai-gray-600}{Dark gray} cells indicate the best result across all models. }
    \label{tab:Base + Pred. Traj(DW) eval}
    \vspace{-0.4cm}
\end{figure*}

\begin{figure*}[t!]
    \captionsetup{type=table}
    \vspace{-0.4cm}
    \centering
    \resizebox{0.9\textwidth}{!}{
    \begin{tabular}{r|cc|ccccccccc}
    & & &
    \rotatebox{75}{Speed Change} &
    \rotatebox{75}{Turn Change} &
    \rotatebox{75}{Lane Change} &
    \rotatebox{75}{Rel. Dist.} & 
    \rotatebox{75}{Rel. Pos.} &
    \rotatebox{75}{Ped. Int.} &
    \rotatebox{75}{E-V Rel. Pos.} &
    \rotatebox{75}{Risk Area} &
    \rotatebox{75}{CFP} \\
    Methods & Rank & Overall & \multicolumn{5}{c}{\cellcolor{red!10}Low-level} & \multicolumn{3}{c}{\cellcolor{orange!10}Mid-level} & \multicolumn{1}{c}
    {\cellcolor{yellow!10}High-level}\\
    \hline
    \rowcolor{navyblue!5}
    \multicolumn{1}{l|}{\textcolor{black}{\textit{Open-source Models}}} & & & & & & & & & & & \\
    Qwen2.5-VL-72B &11 &39.39 &\cellcolor{oai-gray-600}{37.17} &58.17 &83.17 &\cellcolor{oai-gray-600}{36.79} &18.29 &21.74 &16.05 &26.67 &13.25 \\
    Qwen2.5-VL-7B  & 6&42.78 &26.33 &90.67 &96.50 &27.64 &14.43 &37.68 &3.70 &39.33 &8.43 \\
    Qwen3-VL-32B & 5&43.33 &35.17 &78.00 &93.83 &34.35 &14.84 &31.88 &20.99 &53.33 &7.63 \\
    Qwen3-VL-8B & 8&41.40 &29.33 &81.17 &90.17 &32.32 &14.63 &33.33 &11.11 &56.67 &4.55\\
    \hline
    \rowcolor{navyblue!5}
    \multicolumn{1}{l|}{\textcolor{black}{\textit{Closed-source Models}}} & & & & & & & & & & & \\
    GPT-4o Mini & \cellcolor{oai-green-200}{3}&43.98 &34.17 &\cellcolor{oai-gray-600}{92.00} &97.00 &30.49 &14.43 &20.29 &4.94 &6.67 &12.99\\
    GPT-5 &7 &41.71 &27.33 &54.17 &94.33 &31.71 &\cellcolor{oai-gray-600}{25.00} &\cellcolor{oai-gray-600}{40.58} &\cellcolor{oai-gray-600}{48.15} &39.33 &18.47\\
    Gemini-2.0 Flash & 10&39.62 &32.17 &59.67 &32.00 &36.38 &17.48 &27.54 &13.58 &36.67 &8.70 \\
    Gemini-2.5 Flash &9 &40.82 &28.83 &48.33 &95.67 &32.52 &18.29 &27.54 &22.22 &54.67 &\cellcolor{oai-gray-600}{21.15}\\
    Gemini-2.5 Pro & 4&43.51 &32.33 &69.50 &95.17 &30.49 &18.09 &30.43 &28.40 &33.33 &20.35\\
    Claude-3.7-Sonnet &\cellcolor{oai-green-600}{1} &45.58 &29.00 &85.83 &\cellcolor{oai-gray-600}{97.17} &33.74 &16.46 &28.99 &20.99 &37.33 &17.94\\
    Claude-Sonnet-4.5 & \cellcolor{oai-green-400}{2}&44.71&27.83&70.33&97.17&35.16 &19.72 &24.64 &12.35 &\cellcolor{oai-gray-600}{64.00} &19.81\\
    \hline
    \rowcolor{navyblue!5}
    \multicolumn{1}{l|}{\textcolor{black}{\textit{Finetuned Model}}}& & & & & & & & & & & \\
    Qwen3-VL-8B-FI &- &59.54 &39.67 &91.67 &97.17 &42.48 &29.27 &40.58 &41.98 &57.33 &54.75\\
    \end{tabular}
 }
    \caption{\textbf{Baseline \& Pred. V+T (from DrivingWorld) Evaluation on \bench:} Input historical frames, trajectories, predicted frames and trajectories generated from DrivingWorld. \colorbox{oai-gray-600}{Dark gray} cells indicate the best result across all models. }
    \label{tab:Base + Pred. V+T(DW) eval}
    \vspace{-0.4cm}
\end{figure*}

\section{Broader}
The development of Foresight Intelligence in VLMs and WMs holds significant implications for the advancement of embodied AI and autonomous systems. In this section, we discuss the potential positive impacts and ethical considerations associated with our work, \dataset and \bench.

\textbf{Advancing Safety in Autonomous Systems.} The primary motivation behind \dataset is to enhance the safety of autonomous driving. Traditional perception systems often focus on ``what is happening now,'' whereas safe driving necessitates anticipating ``what will happen next.'' By fostering research into Foresight Intelligence, specifically capabilities like Risk Area Assessment and Counterfactual Prediction, our work encourages the development of models that can proactively identify hazards and reason about ``what-if'' scenarios. This shift from reactive perception to proactive anticipation is a critical step toward reducing accidents and 
{supporting research toward safer autonomous-driving systems.}

\textbf{Ensuring Trustworthiness in Generative World Models.} With the rapid rise of generative AI, World Models are increasingly used to simulate future driving scenarios. However, a major risk is that these models may generate visually realistic but physically implausible or unsafe futures. \bench introduces a mechanism to evaluate the semantic coherence of generated data. By rigorously testing whether generated futures align with causal logic and safety rules, our benchmark helps prevent the deployment of unreliable simulators and promotes the development of trustworthy World Models for safety-critical applications.

\textbf{Limitations and Fairness.} We acknowledge that the data in \dataset, inherited from nuScenes, is geographically limited to specific urban environments (Boston and Singapore). Driving behaviors, traffic rules, and road infrastructures vary significantly across different countries and cultures. Consequently, models trained or evaluated on this benchmark may exhibit geographic bias and may not generalize perfectly to rural areas or developing regions with different traffic dynamics. We urge researchers and practitioners to exercise caution and perform extensive testing before deploying such models in unseen domains.

\section{More Qualitative Visualization. }
\begin{figure*}[t]
  \centering
   \includegraphics[width=0.95\linewidth]{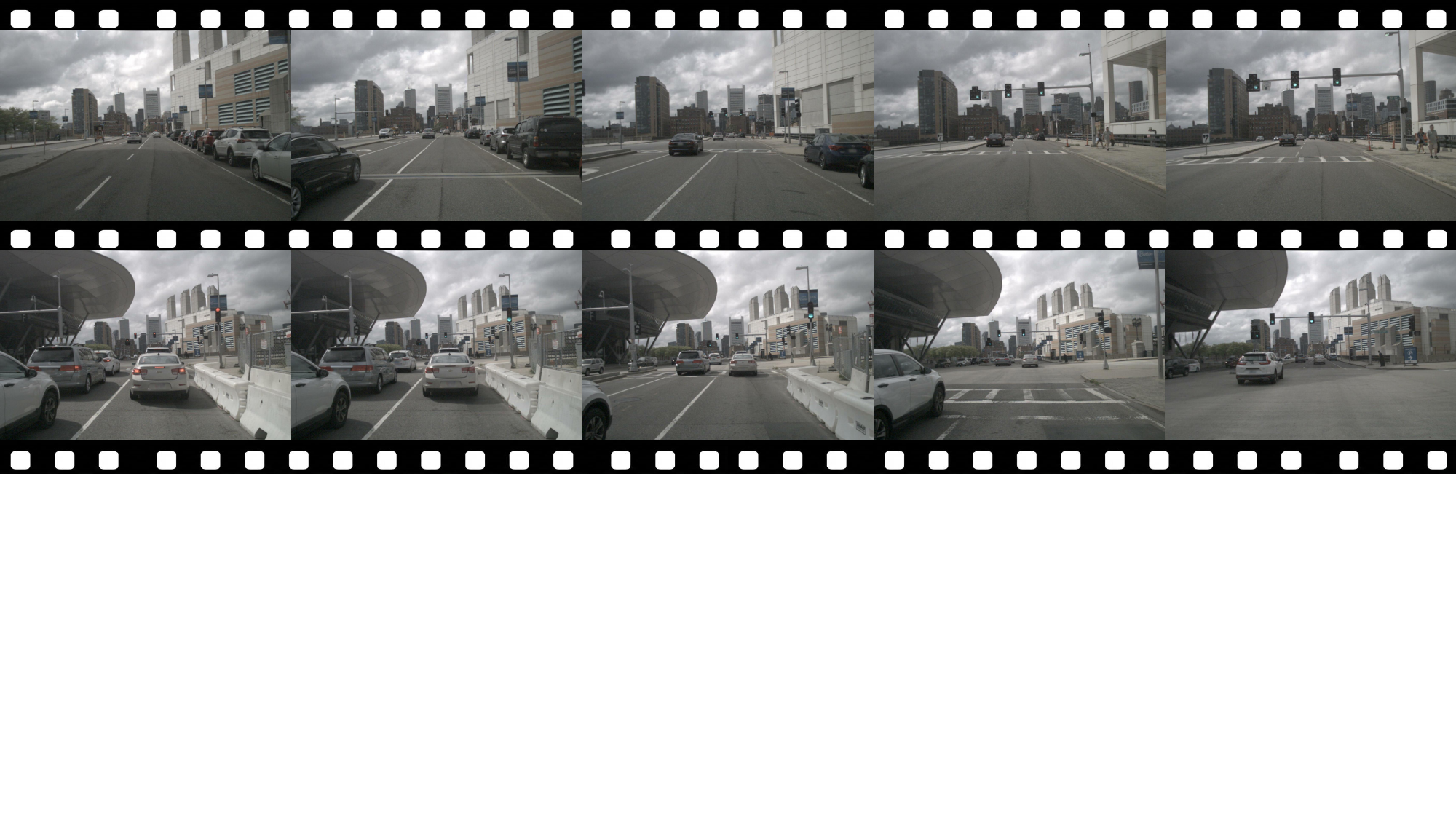}
   \caption{\textbf{Speed Change Visualization. Top:} Constant Speed. \textbf{Bottom:} Acceleration.}
   \label{fig:speed change}
\end{figure*}

\begin{figure*}[t]
  \centering
   \includegraphics[width=0.95\linewidth]{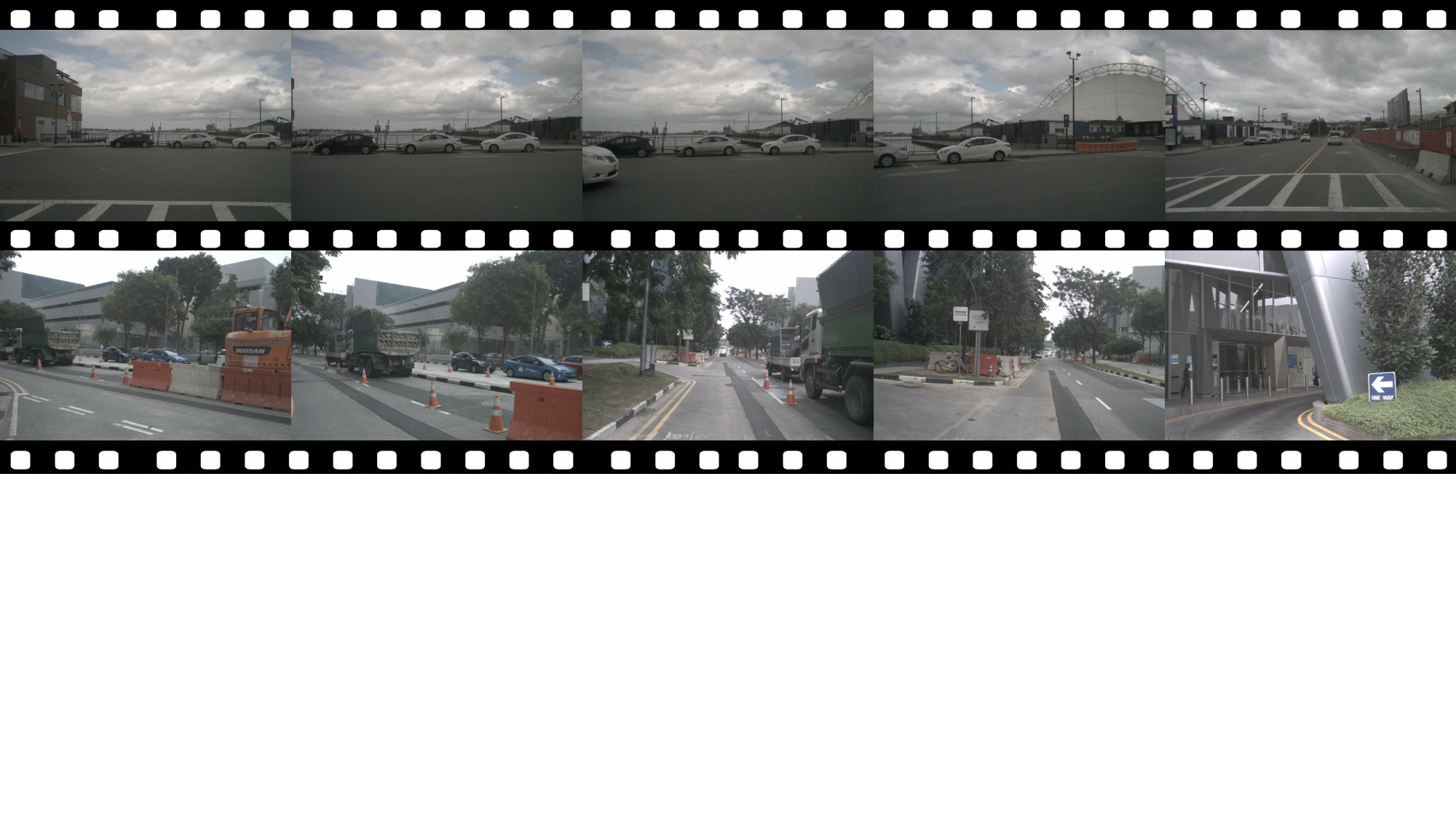}
   \caption{\textbf{Turn Change Visualization. Top:} Turn Right. \textbf{Bottom:} Turn Left.}
   \label{fig:turn change}
\end{figure*}

\begin{figure*}[t]
  \centering
   \includegraphics[width=0.95\linewidth]{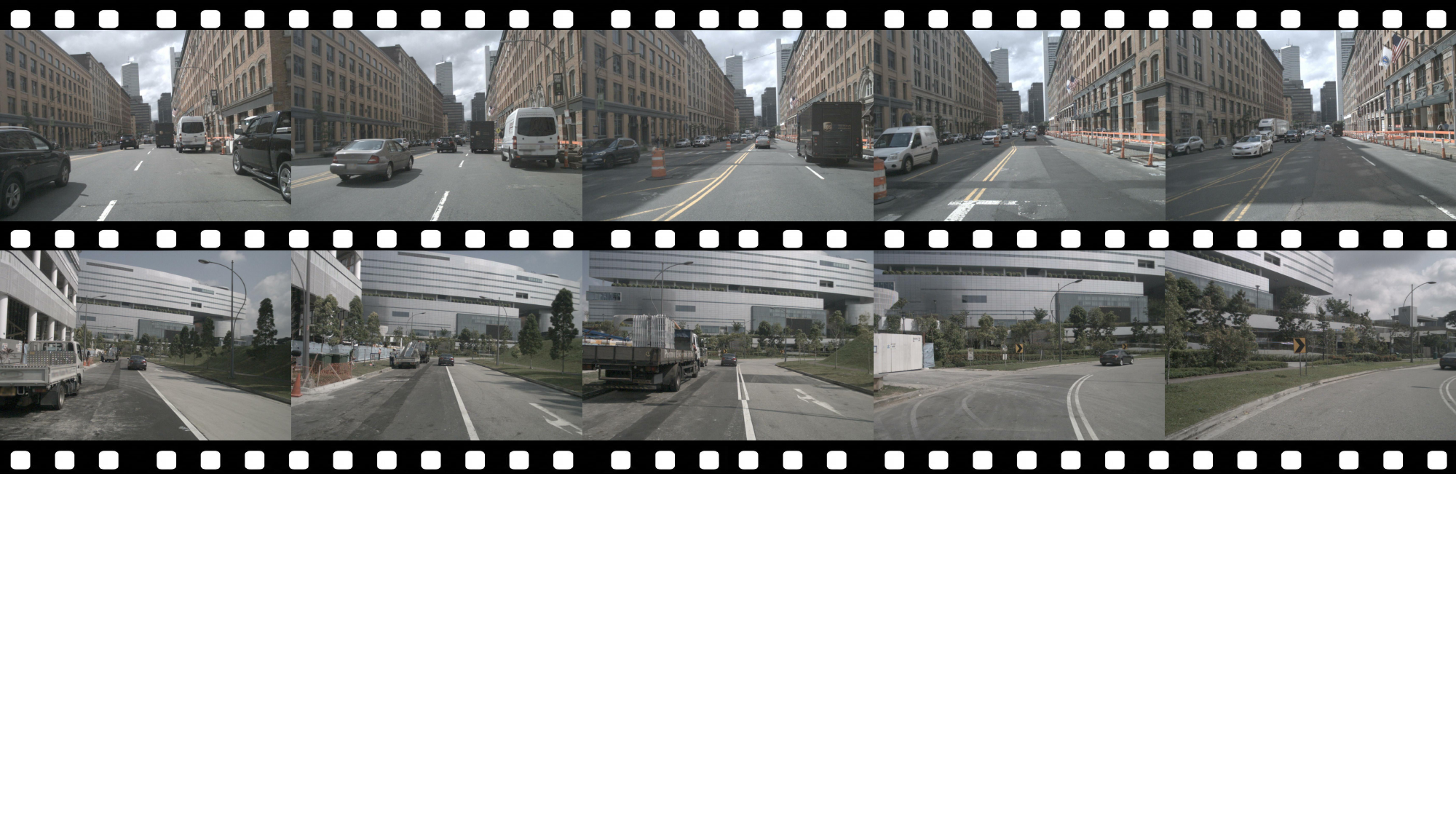}
   \caption{\textbf{Lane Change Visualization. Top:} Left Lane Change. \textbf{Bottom:} Right Lane Change.}
   \label{fig:Lane change}
\end{figure*}

\begin{figure*}[t]
  \centering
   \includegraphics[width=0.95\linewidth]{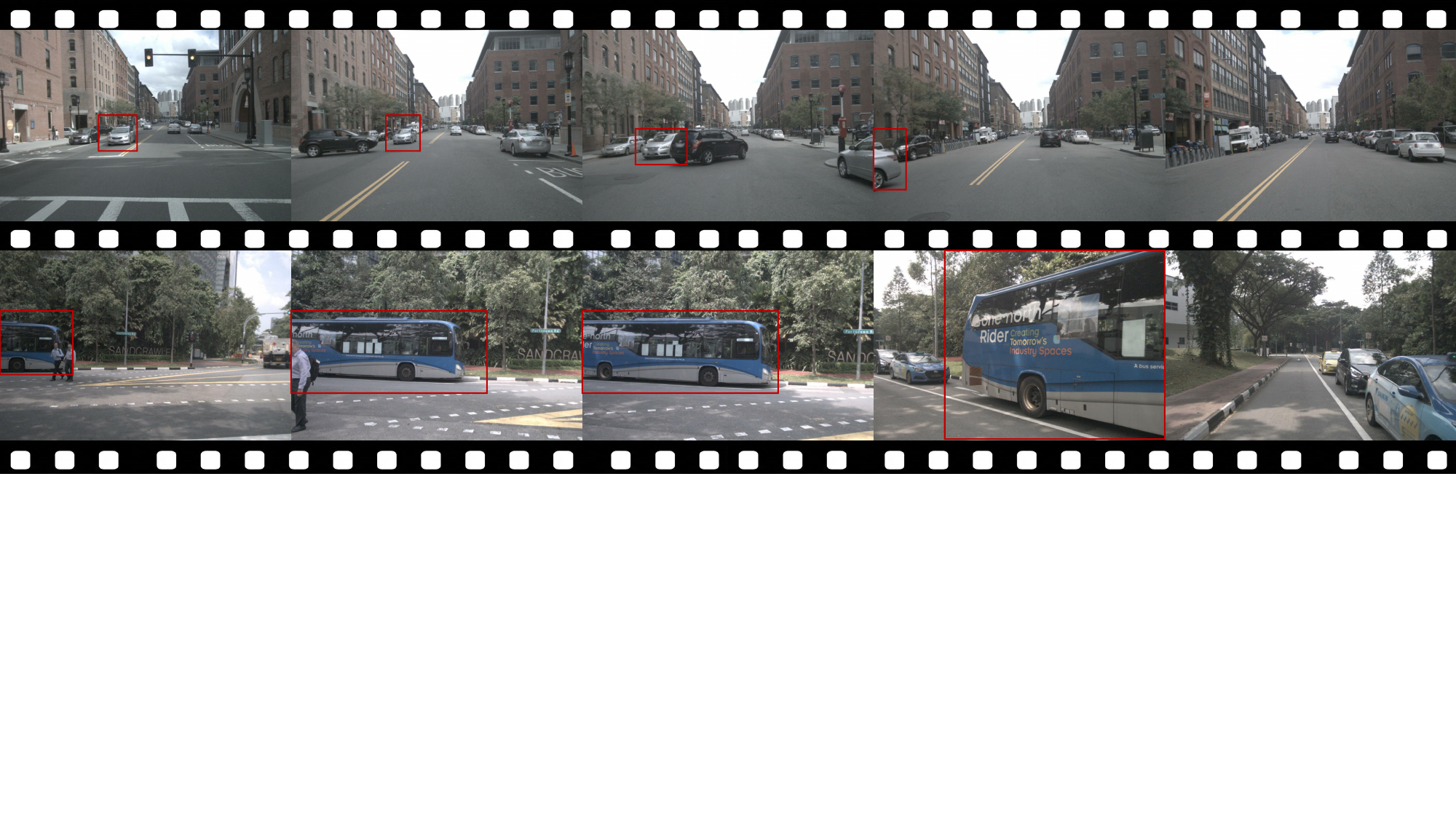}
   \caption{\textbf{Relative Distance \& Relative Position Visualization.} The bounding
boxes are illustrative only and do not appear in the actual dataset.}
   \label{fig:Relative}
\end{figure*}

\begin{figure*}[t]
  \centering
   \includegraphics[width=0.95\linewidth]{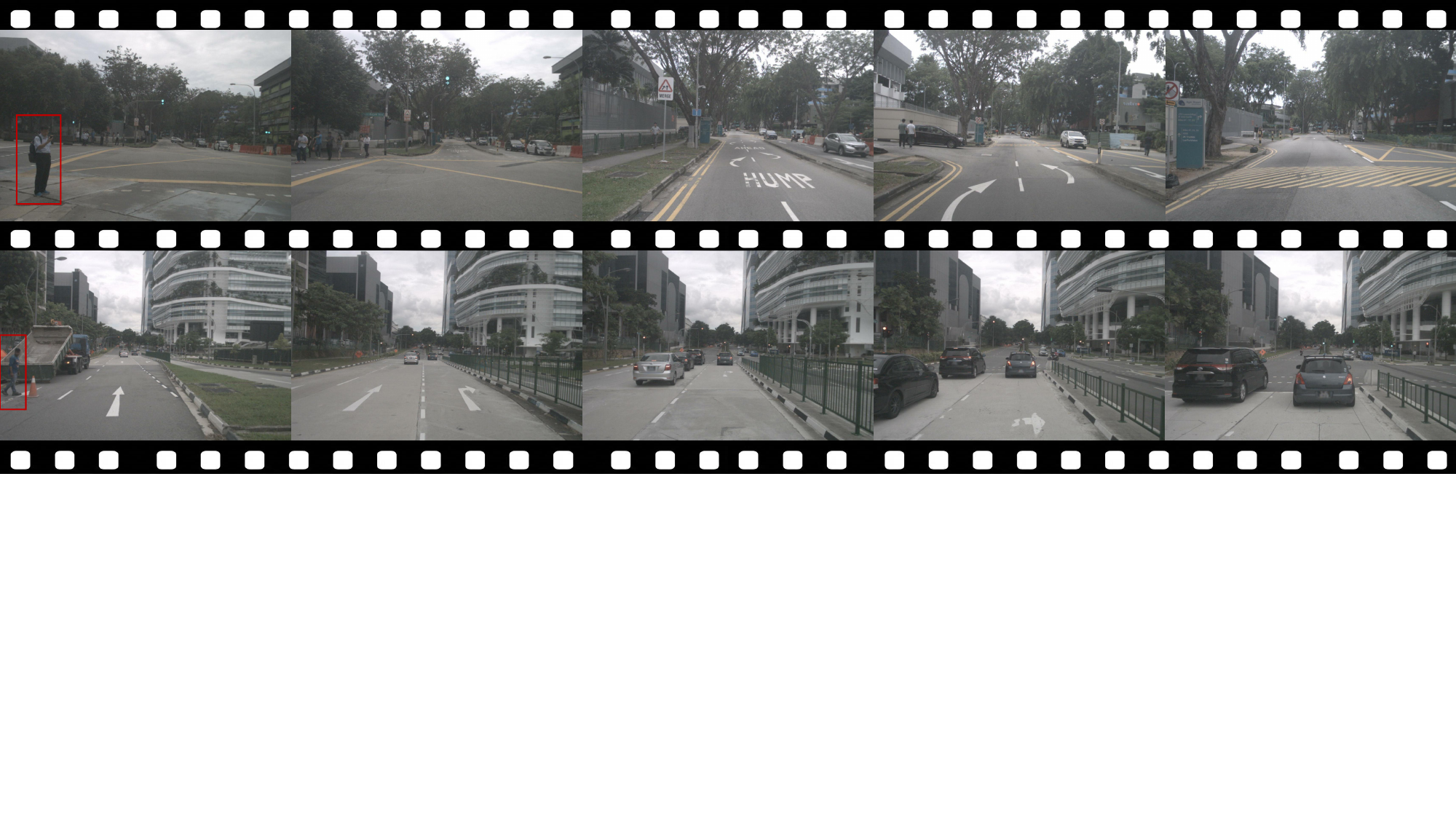}
   \caption{\textbf{Pedestrian Intent Visualization. Top:} Waiting at curbside. \textbf{Bottom:} Jaywalking. The bounding
boxes are illustrative only and do not appear in the actual dataset.}
   \label{fig:Pedestrian}
\end{figure*}

\begin{figure*}[t]
  \centering
   \includegraphics[width=0.95\linewidth]{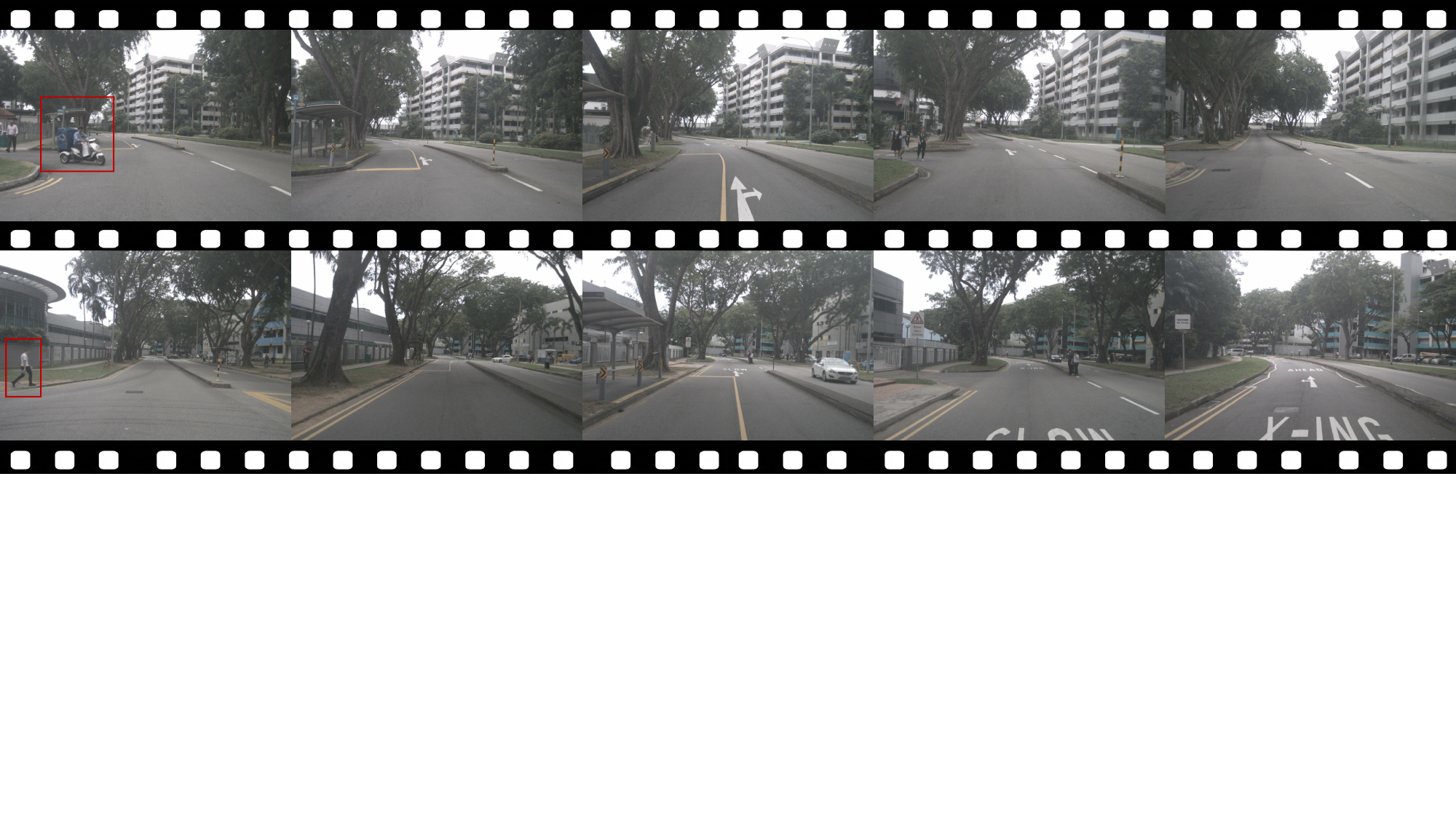}
   \caption{\textbf{Ego-VRU Relative Position Visualization. Top:} Motorcyclist. \textbf{Bottom:} Pedestrian. The bounding
boxes are illustrative only and do not appear in the actual dataset.}
   \label{fig:Ego-VRU}
\end{figure*}

\begin{figure*}[t]
  \centering
   \includegraphics[width=0.95\linewidth]{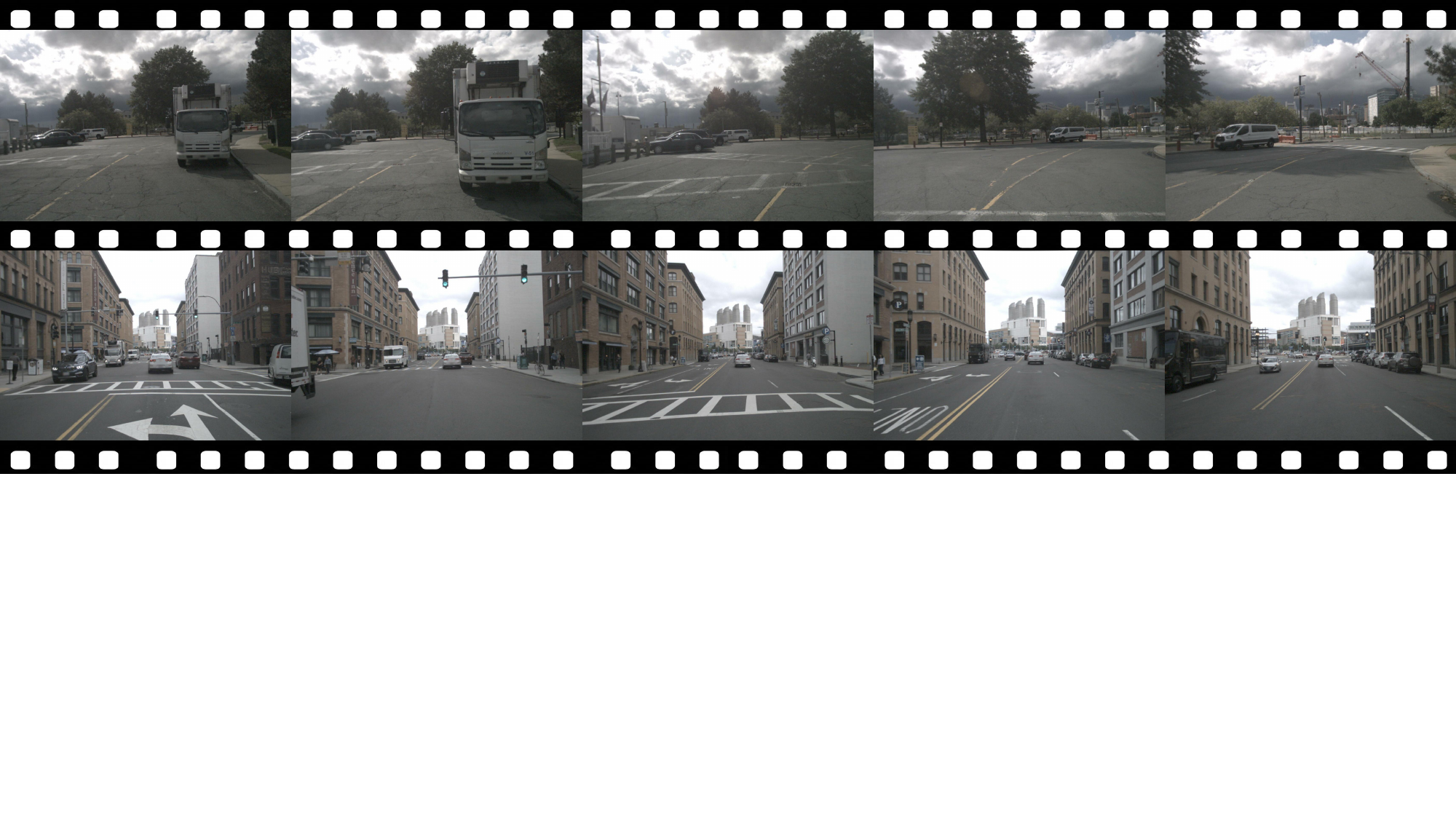}
   \caption{\textbf{Risk Area Visualization. Top:} High Occlusion Area. \textbf{Bottom:} Complex Intersection.}
   \label{fig:risk_area}
\end{figure*}

\begin{figure*}[t]
  \centering
   \includegraphics[width=0.95\linewidth]{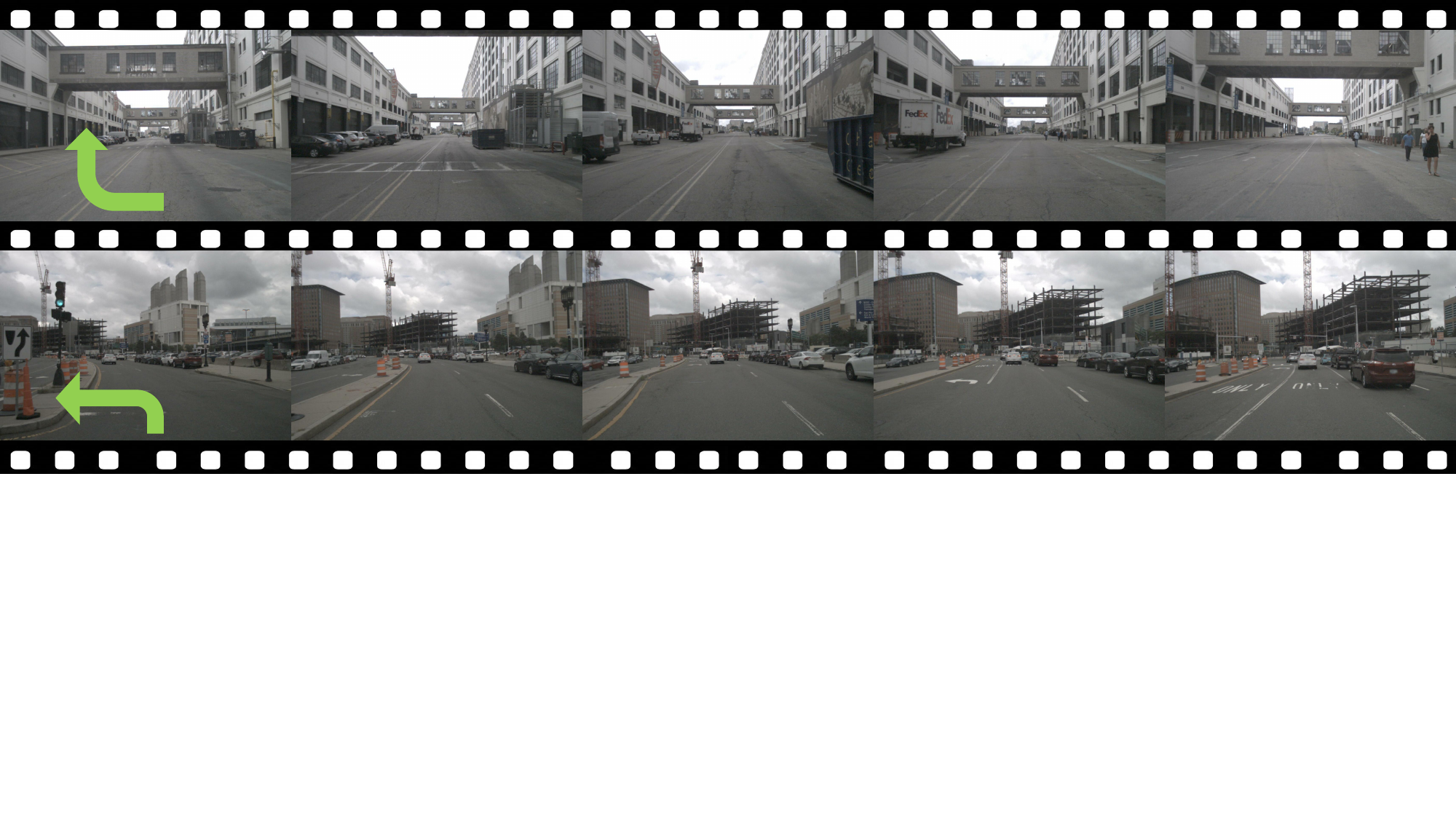}
   \caption{\textbf{Counterfactual Prediction Visualization. Top:} What would happen if the ego vehicle were to change lanes to the left? \textbf{Bottom:} What would happen if the ego vehicle were to turn right?}
   \label{fig:cfp}
\end{figure*}

Visualizations for the nine task types are illustrated in \cref{fig:speed change,fig:turn change,fig:Lane change,fig:Relative,fig:Pedestrian,fig:Ego-VRU,fig:risk_area,fig:cfp}.

\end{document}